\documentclass[lettersize,journal]{IEEEtran}
\usepackage{amsmath,amsfonts}
\usepackage{algorithmic}
\usepackage{algorithm}
\usepackage{array}
\usepackage[caption=false,font=normalsize,labelfont=sf,textfont=sf]{subfig}
\usepackage{textcomp}
\usepackage{stfloats}
\usepackage{url}
\usepackage{verbatim}
\usepackage{balance}
\usepackage{stfloats}
\usepackage{graphicx}
\usepackage{cite}
\usepackage{booktabs}
\usepackage{multirow}
\usepackage{enumitem}
\usepackage{subfig}
\begin{document}

\title{Multiple Instance Learning for Cheating Detection and Localization in Online Examinations}

\author{Yemeng Liu, Jing Ren, Jianshuo Xu, Xiaomei Bai, Roopdeep Kaur, and Feng Xia,~\IEEEmembership{Senior Member,~IEEE}
\thanks{Y. Liu and J. Xu are with School of Software, Dalian University of Technology, Dalian 116620, China. Email: liuziweiww@outlook.com, xujianshuo0243@gmail.com.}
\thanks{J. Ren, and F. Xia are with School of Computing Technologies, RMIT University, Melbourne, VIC 3000, Australia. Email: ch.yum@outlook.com, f.xia@ieee.org}
\thanks{X. Bai is with School of Artificial Intelligence, Anshan Normal University, Anshan 114007, China. Email: xiaomeibai@outlook.com}
\thanks{R. Kaur is with Institute of Innovation, Science and Sustainability, Federation University Australia, Ballarat, VIC 3353, Australia. Email: roopdeepkaur@students.federation.edu.au}
\thanks{Corresponding author: Feng Xia}

}

\markboth{IEEE Transactions on Cognitive and Developmental Systems, Dec~2023}%
{Yemeng \MakeLowercase{\textit{et al.}}: A Sample Article Using IEEEtran.cls for IEEE Journals}

\IEEEpubid{}

\maketitle

\begin{abstract}
The spread of the Coronavirus disease-2019 epidemic has caused many courses and exams to be conducted online. The cheating behavior detection model in examination invigilation systems plays a pivotal role in guaranteeing the equality of long-distance examinations. However, cheating behavior is rare, and most researchers do not comprehensively take into account features such as head posture, gaze angle, body posture, and background information in the task of cheating behavior detection. In this paper, we develop and present CHEESE, a CHEating detection framework via multiplE inStancE learning. The framework consists of a label generator that implements weak supervision and a feature encoder to learn discriminative features. In addition, the framework combines body posture and background features extracted by 3D convolution with eye gaze, head posture and facial features captured by OpenFace 2.0. These features are fed into the spatio-temporal graph module by stitching to analyze the spatio-temporal changes in video clips to detect the cheating behaviors. Our experiments on three datasets, UCF-Crime, ShanghaiTech and Online Exam Proctoring (OEP), prove the effectiveness of our method as compared to the state-of-the-art approaches, and obtain the frame-level AUC score of 87.58\% on the OEP dataset.
\end{abstract}


\section{Introduction}\label{sec1}
\begin{figure*}[htbp]
\centering
\includegraphics[scale=0.365]{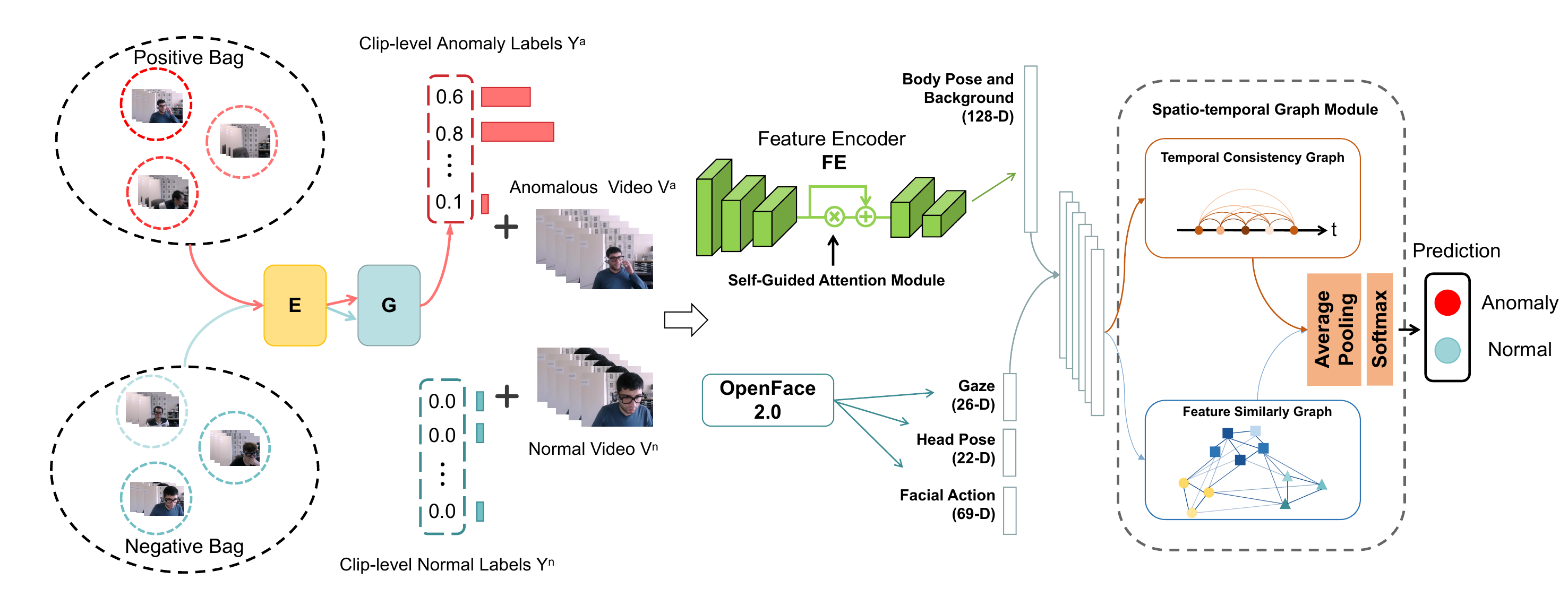}
\caption{The flow chart of the proposed CHEESE which consists of a multiple instance label generator $G$ and a feature encoder $FE$ followed by a spatio-temporal graph module. We utilize feature extractor $E$ to provide clip-level features for the label generator and apply the clip-level labels  $ Y^a =\left \{y_i^a\right \}$ and $Y^n$ to train the feature encoder in the second stage. Specifically, the positive and negative bags are divided according to the video-level label $Y=0/1$. $y_i^a$ is generated by the generator, and $Y^n$ can be directly derived from the video-level label ($Y=0$).}
\label{fig_1}
\end{figure*}

\IEEEPARstart{D}{ue} to the newly prevailed Coronavirus disease-2019 (COVID-19), online examinations have gradually become an indispensable means of assessing students' knowledge standard  \cite{guo2021educational}. Many scholars and service providers are making efforts to ensure the equality of online examinations. How to detect cheating behaviors during examinations has become an issue of great concern.

Video anomaly detection \cite{liu2018future, ionescu2019object,tudor2017unmasking,nguyen2019anomaly,fang2020anomaly,lu2013abnormal} generally refers to the process of identifying the event that significantly deviates from the normal behaviors. The detection and location of abnormal behaviors, such as stampedes, traffic accidents, or terrorist attacks, play an important role in the security and protection system. However, manual annotation is time-consuming when it comes to various abnormal events. Moreover, collecting anomalous datasets is a difficult task due to the rare occurrence of anomalous events in real life. Therefore, video anomaly detection is generally regarded as an unsupervised learning task in previous studies \cite{zhao2017spatio,luo2017revisit,sultani2018real,gong2019memorizing,zhou2019anomalynet}, which only utilizes normal training samples to learn the normal patterns.

The process of abnormal behaviors detection is to identify the difference between the learned normal and abnormal behavior feature representations. Different from traditional anomaly detection tasks where cameras could provide a broad and full-scaled view, cheating detection has much more complicated and subtle challenges. In most online examinations, the examinees are asked to turn on their webcams and the examiners can only see a limited view, which is usually composed of the body above the chest and the surroundings around the examinees. Considering that the picture information provided by the webcam is very limited, and the seriousness of the misjudgment of cheating in different countries, the applicable scenarios of the algorithm need to be strictly regulated.

To solve these problems, several models have been proposed to detect cheating behaviors from different perspectives. Recently, Multiple Instance Learning (MIL) \cite{sultani2018real,nguyen2018weakly} has gradually become the main technique for anomaly detection tasks. It treats the video as a package, and uses the splitted clips of each video as an instance. Then, the model tries to find out the clips of abnormal events in the abnormal video. However, MIL can only locate anomalies in the temporal dimension, while ignoring other discriminative features. For example, face recognition technology \cite{sun20223d} is used to grab the number of human faces in a certain time period through a webcam to determine whether there is an external assistant in the examination room. In addition, the object detection algorithm \cite{ionescu2019object} is also used to identify and analyze the existing items in the streaming media. With the help of the gaze estimation algorithm, \cite{cheung2015eye}, the examinee's cheating behavior patterns can be identified by calculating the gaze range.

In previous works \cite{zhong2019graph,chang2021contrastive,wang2018videos,markovitz2020graph,li2020peer}, Graph Convolutional Network (GCN) has been used in anomaly detection because of its ability to effectively capture dependencies between clips. In the graph, clips are abstracted as vertices, and abnormal information is propagated through the edges. Further, feature similarity \cite{zhong2019graph} and temporal consistency \cite{chang2021contrastive,tian2021weakly} are used as two features in GCN to clean the noise of the label. Feature similarity means that the abnormal clips have some similar features, while temporal consistency indicates that the abnormal clips may be close to each other in the time dimension. In real cheating scenarios, the candidate's head and surrounding scenes tend to be abnormal for a specific short period of time, which means that the spatio-temporal features in the video are very important. Moreover, the ambiguous boundary between normal and abnormal data impedes the performance of unsupervised learning models. This work is dedicated to solving cheating problems in online closed-book test scenarios. In this type of exam scenario, scratch paper and calculators will be provided in the form of electronic tools, and candidates are not allowed to carry books, mobile phones, and other items without authorization. To solve the cheating problems in this scenario, we propose a method to capture anomalies in the video frames in a weakly supervised way. The videos are annotated as abnormal if it contains abnormal frames, while the temporal information is not provided.

In this work, we apply the clip-level labels generated by multi-instance learning as the supervision of the feature encoder. On this basis, multi-modal features are obtained by fusing GAP features \cite{niu2018automatic}, body pose, and background features for cheating behavior detection. To take advantage of spatio-temporal dependencies between multi-modal features, we introduce a spatio-temporal graph module with temporal consistency and feature similarity. Specifically, we extract the C3D/I3D features from the feature extractor and put them into the label generator. The generator based on MIL can produce clip-level labels for the feature encoder training. In the second stage, through these labels and their corresponding video information, we acquire multi-modal features with the help of openface 2.0 and train our feature encoder with spatio-temporal graph module for learning discriminative features (as shown in Figure~\ref{fig_1}).

To the best of our knowledge, this paper is the first to use a MIL-based approach and comprehensively consider multi-modal features such as facial appearance, head posture, and gaze angle to detect cheating behaviors. 
The contributions of this paper are summarized as follows:
\begin{itemize}[leftmargin=0em,itemindent=2em]
\item{We propose a novel weakly supervised framework for cheating behaviors detection combined with multi-modal features for graph learning. }
\item{We introduce the spatio-temporal graph module to capture the spatio-temporal relationship between clips and propagate the supervision signal, in which the stitched features are comprehensively taken into account. }
\item{We conduct experiments on three different anomaly detection datasets including an online examination monitoring dataset. The experimental results demonstrate the effectiveness of our model.}
\end{itemize}

The remainder of this paper is organized as follows. The related works are discussed in Section~\ref{sec2}. Section \ref{sec3} defines the problem of video anomaly detection. Section \ref{sec4} presents the overall framework of the proposed model and describes the design of each module. The comparative and ablation experiments results will be discussed in Section \ref{sec5} and followed by the conclusion in Section \ref{sec6}.

\section{Related Work}\label{sec2}
Due to the powerful capabilities of deep neural networks in learning informative representations of image data, a large number of deep anomaly detection methods, especially unsupervised learning models, have been introduced for coping with anomaly detection challenges in different real-world applications \cite{pang2021deep}. 

\subsection{Video Anomaly Detection}
As one of the most challenging problems in computer vision, video anomaly detection has been extensively studied for many years. Generally, reconstruction model and predictive methods are used to verify the normality of the model~\cite{ren2021deep,zhou2019anomalynet,zhao2011online,zhang2021video,perera2019ocgan,zhai2016deep,sabokrou2015real,pang2020self,park2020learning,xu2017detecting,ruff2018deep,tian2021weakly}. Reconstruction methods, such as Generative Adversarial Networks (GAN) \cite{zhang2021video,perera2019ocgan}, and Autoencoder \cite{zhai2016deep,sabokrou2015real,pang2020self} attempt to distinguish abnormal events by redrawing the input. In particular, the memory module \cite{park2020learning} is introduced to record the prototype of the normal data, which represents the different patterns of the normal video frames for unsupervised anomaly detection. Predictive methods attempt to encode normal events through statistical models. Xu et al. \cite{xu2017detecting} used Support Vector Machines (SVM) to detect anomalies after extracting the features. Ruff et al. \cite{ruff2018deep} utilized Convolutional Neural Networks (CNNs) to map the normal data as the center of the sphere and exclude the abnormal samples outside it. Tian et al. \cite{tian2021weakly} proposed Multiscale Temporal Network (MTN) to capture the local and global temporal dependencies between video clips, and achieved state-of-the-art results. Inspired by the temporal dependencies of MTN networks, we introduce a framework to capture dependencies between clips: we first optimize the feature encoder in a fine-grained manner by introducing MIL. Then, in order to capture spatio-temporal dependencies between video clips, the spatio-temporal graph module is applied.

\subsection{Online Proctoring}
The methods of online proctoring can be divided into three types: online human proctoring, semi-automatic proctoring and fully automatic proctoring. Online human proctoring means that remote proctors supervise candidates throughout the online exam process. However, this is labor-intensive and the cost increases with the number of students. To eliminate the human cost, fully automatic supervised methods have been proposed, which usually exploit machine learning techniques to identify cheating behaviors. However, existing methods of fully automated proctoring suffer from several problems, including the "black-box" nature of deep learning algorithms and unreliable decision-making caused by biased training datasets. Therefore, it is almost impossible to rely entirely on fully automated methods to determine whether a student is cheating during an online exam. To solve the problems caused by fully automated proctoring methods, semi-automated proctoring was introduced, which allows humans to complete the final decision with the aid of machine learning methods \cite{costagliola2008monitoring,li2015massive,atoum2017automated}. A representative work is the Massive Open Online Processor proposed by \cite{li2015massive}. Their method is the first to use machine learning techniques to detect suspected student cheating and send the detections to teachers for further review. However, it requires each candidate to use multiple devices for online exams, which increases exam costs. Costagliola et al.  \cite{costagliola2008monitoring} proposed a visual analysis system to assist proctors in invigilating an exam, but it is limited to identifying the cheating behavior that a student is looking at another student's screen, which is not enough in online exams. Atoum et al. \cite{atoum2017automated} developed a multimedia analysis system to detect various cheating behaviors during online exams. But their approach requires two cameras which is unrealistic contemporarily. Li et al. \cite{li2021visual} facilitate the supervision of online exams by analyzing video recordings of exams and mouse movement data from each student. In addition, the angles of the head posture have also become the key point to determine whether the examinee is cheating \cite{yang2019fsa,kuhnke2019deep,ruiz2018fine}. But most researchers tend to focus on a few features, without fully considering head posture, gaze angle, body posture, background, and other features.

Different from the methods above, we use the weakly supervised method to identify the above three types of cheating behaviors (not include the types of cheating behavior that can be detected by the system level, such as obtaining page monitoring, and voice access for direct detection). In addition, a variety of behavioral features, including head posture, gaze angle, body posture, background, and facial appearance, are considered comprehensively for cheating detection.
In summary, our method is able to efficiently detect candidate cheating behaviors while ensuring real-time performance.

\subsection{Multiple Instance Learning}
In recent years, MIL has shown remarkable performance in the field of video anomaly detection. MIL is a weakly supervised learning method which treats the video as a package, and uses the splitted clips of each video as an instance \cite{ruff2018deep,nguyen2018weakly}. With the help of a specific feature aggregation function, video-level annotations can be used to supervise learning in the instance level. MIL serves as an anomaly classifier, using the features extracted from the action classifier to detect anomalies, and deep MIL ranking loss to calculate anomaly scores. On this basis, Wan et al. \cite{wan2020weakly} introduced dynamic loss and center-guided regularization. Similarly, Zhu et al. \cite{zhu2019motion} proposed an attention-based MIL model, and introduced an optical flow based autoencoder for feature encoding.

Different from them, we use MIL as a label generator for training, and
learn discriminative features in the video frames by adding a feature encoder.
In particular, Feng et al. \cite{feng2021mist} used a similar approach as ours to build the model. In contrast, we propose the spatio-temporal graph module in the feature encoder to capture the spatio-temporal dependency information between clips. Further, in the second stage, we use the multi-modal features, which allow the spatio-temporal graph module to comprehensively consider the relationships between the feature of clips.

\subsection{Graph Convolutional Neural Network}
The GCN \cite{kipf2016semi,ou2016asymmetric,pfeiffer2014attributed,liu2022deep,xia2021graph} has been extensively studied on the problem of processing graph-structured data. In the field of video anomaly detection, GCN is often used to mine various relationships between features \cite{zhong2019graph, chang2021contrastive}. Zhong et al. \cite{zhong2019graph} applied GCN for the first time to clean the label noise in anomaly detection datasets, and exploited two relationships of feature, namely feature similarity and temporal consistency, to correct label noise. Similarly, Chang et al. \cite{chang2021contrastive} proposed a contrastive attention module to address the problem of abnormal data imbalance, which mainly consists of a temporal graph convolutional layer to utilize video temporal context.

In the process of cheating behaviors detection, the feature similarity graph can effectively capture the long-term relationship between objects/regions and promote the propagation of supervised signals of clip similarity \cite{zhong2019graph}. In addition, the cheating behaviors of candidates may appear at close time points, which conforms to the characteristic of temporal consistency. Therefore, we establish a spatio-temporal graph in the graph module of the feature encoder to obtain the spatio-temporal relationship between clips, thereby improving the prediction performance of the encoder.

\section{Problem Definition}\label{sec3}
\textbf{Cheating behaviors} may occur at any one stage of the examination process. The following events are considered during an online exam: \textit{Another Person, Absence, and Device}. \textit{Another Person} is a situation where the assistant is off-screen, or where parts of the helper's face or body appear on the screen. In the case of the off-screen assistants, the assistant generally transmits the answer through voice. Obtaining the candidate's voice authority by the browser can effectively stop this type of cheating behavior, because the candidate is not allowed to make a sound in the online exam. But when only part of the assistant appears on the screen and whispers up close, the system has difficulty catching cheating through voice. In this case, The footage captured by the webcam is entered into our method for cheating detection. Similar to \textit{Another Person}, \textit{Device} is also divided into two situations: off-screen and on-screen. But in either case, the candidate's head will be abnormal for a period of time (e.g., bowing head, turning head sideways). Thus, fine-grained head pose features can be used as important information for algorithms to judge cheating behaviors. Furthermore, the videos will be labelled as \textit{Absence} if the camera fails to detect the examinee’s face or body.

In addition to the above three types of events, candidates can also search for answers on the same computer through a search engine. This situation is solved in the same way as the off-screen helper, page or voice monitoring permissions are obtained by the browser for the detection of both types of events. When the voice reaches a threshold or a candidate exits the exam page, the proctor system automatically raises an alert. 

\textbf{Multiple Instance Learning}: Given a video $V$$=$$\{v_i\}^N_{i=1}$ that can be divided into $N$ clips, it can be known whether the video contains abnormal clips through the video-level label $Y$$\in$$\{1,0\}$. In other words, the training data did not contain instance-level annotations. Anomaly detection models need to accurately locate the timestamp of the abnormal event in the video. This kind of video anomaly detection problem under a weak supervisory signal is regarded as a typical multiple instance learning problem. For better understanding, we provide some MIL-related definitions:

MIL is a kind of supervised learning. The training set $X=\{x_1,x_2,...,x_n\}$ is composed of $n$ bags, and each bag $x_i=\{o_1,o_2,...,o_m\}$ contains $m$ instances. A bag $x_i$ is labeled positive ($y_i=1$) if it contains at least one positive instance, while in a negative bag ($y_i=0$), all instances should be negative. The task of MIL is to learn informative concepts from the training set and locate the position of the positive instance in the positive bag. The label $y_i$ of bag $x_i$ is defined as:
\begin{equation}
y_i=\left\{
\begin{array}{rcl}
1 & & {if \exists o_1 : o_m = 1}\\
0 & & {if \forall o_1 : o_m = 0}
\end{array} \right.
\end{equation}

\textbf{MIL in Video Anomaly Detection} treats the video $V$ as a bag, and the clip $v_i$  as an instance. Specifically, the negative bag is denoted as $B^n$$=$$\left \{ {v^{n}_i}\right \}^N_{i=1}$, where each clip/instance $v^{n}_i$ in the video is labeled negative ($Y$$=$$0$); while the positive bag $B^a$$=$$\left \{ {v^{a}_i}\right \}^N_{i=1}$ contains at least one anomalous instance ($Y$$=$$1$).

\textbf{Problem Definition}: The video anomaly detection problem is modeled as an instance detection problem under multiple instance learning. By calculating the anomaly score of each instance $v^a_i$ in the positive bags, anomaly clips in the video which are not annotated can be detected from a weak monitoring signal.

\begin{algorithm}[H]
\caption{Pseudo code of CHEESE.}\label{alg:alg1}
\begin{algorithmic}
\STATE  \textbf{Input:}\hspace{0.1cm}Video-level labeled videos $V$, corresponding video-level labels $Y$ and feature extractor $E$.
\STATE  \textbf{Output:}\hspace{0.1cm}Clip-level Anomaly labels  $ Y^a =\left \{y_i^a\right \}^N_{i=1}$  for anomaly clips $\left \{ {v^{a}_i}\right \}^N_{i=1}$  in stage I and predicted labels L in stage II.
\STATE $ \textbf{Stage I: Labels Generation}  $
\STATE \hspace{0.5cm}1: Split the video into clips $\{v_i\}^N_{i=1}$ as input to the feature extractor $E$
\STATE \hspace{0.5cm}2: Extract clip features $ \left \{ {f_i}\right \}^N_{i=1} $ of $\{v_i\}^N_{i=1}$ from $E$.
\STATE \hspace{0.5cm}3: Train the label generator $G$ with $ \left \{ {f^{n}_i}\right \}^N_{i=1} $ and their corresponding video-level labels.
\STATE \hspace{0.5cm}4: Generate clip-level anomaly labels $\left \{ {y^{a}_i}\right \}^N_{i=1}$ for each anomaly clips $\left \{ {v^{a}_i}\right \}^N_{i=1}$.
\STATE // $ Y^a $ are mixed with $Y^n$ (derived from $Y=0$) as clip-level labels of feature encoder.
\STATE $ \textbf{Stage II: Train the Feature Encoder}  $
\STATE \hspace{0.5cm}1: Place the clips $\{v_i\}^N_{i=1}$ into openFace 2.0 and feature encoder to extract features separately.
\STATE \hspace{0.5cm}2: Combine different features into a complete feature.
\STATE \hspace{0.5cm}3: Convert stitched features into the spatio-temporal graph.
\STATE \hspace{0.5cm}4: Train the feature encoder under the supervision of $Y^n$ and $Y^a$.
\end{algorithmic}
\label{alg1}
\end{algorithm}

\begin{figure}[htbp]
	\centering
	\includegraphics[scale=0.4]{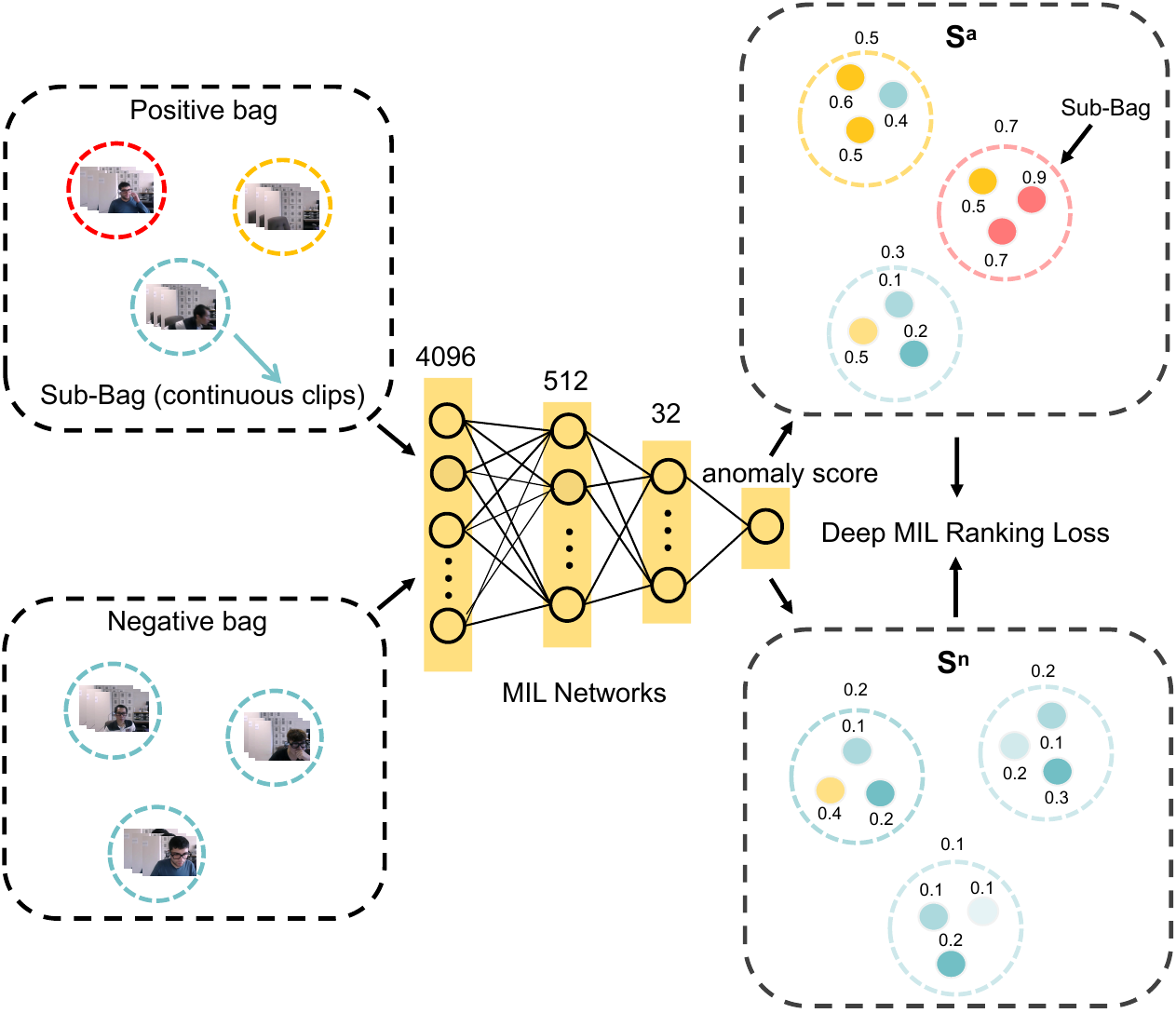}
	\caption{The structure of our label generator. After the features are given, we exploit continuous sampling to obtain sub-bags. Each sub-bag contains the features of T consecutive clips. }
	\label{fig2}
\end{figure}

\section{Methodology}\label{sec4}
In this section, we will introduce the proposed model CHEESE shown in Figure \ref{fig_1}. Specifically, CHEESE is mainly composed of two parts: the label generator and the feature encoder (as shown in Algorithm \ref{alg1}).

\subsection{Multiple Instance Learning for Label Generation}

\begin{figure*}[htbp]
	\centering
	\includegraphics[scale=0.35]{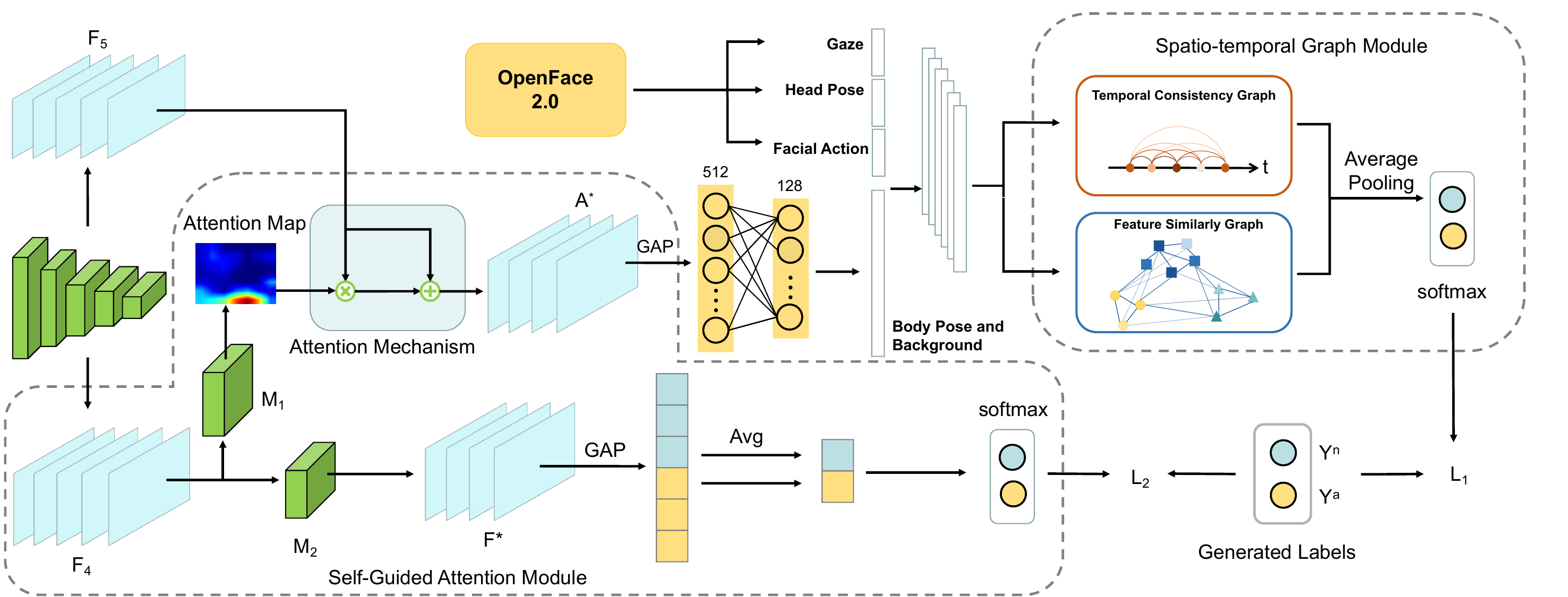}
	\caption{The structure of our dual branch attention enhanced feature encoder. $F_4$, $F_5$, $F^*$ and $A^*$ are characteristic maps. $M_1$ and $M_2$ are two coding modules constructed by convolutional layer. $L_1$ and $L_2$ are cross entropy loss functions. GAP is the global average pooling operation, and Avg represents the operation of channel-level average pooling. }
	\label{fig3}
\end{figure*}

According to the characteristics of multiple instance learning, each instance in the negative bag is negative, while the instances in each positive bag are unidentified. Therefore, we need to use the label generator to generate clip-level anomaly scores for the instances in the positive bags as their labels, and differentiate the anomaly scores between positive and negative instances as accurately as possible. It should be noted that a clip is usually composed of several frames, and the number of frames in a clip could be adjusted according to the experimental results. In the feature extraction stage, we deploy a classical feature encoder, i.e. C3D \cite{tran2015learning} pretrained on Sport-1M \cite{karpathy2014large} or I3D pretrained on Kinetics-400 \cite{carreira2017quo} to extract clip-level features required by the label generator. However, if the video of indeterminate length is simply fused into 32 segment features as the input of the multilayer perceptron (MLP) \cite{sultani2018real}, the prediction effect of the final model is often not ideal. In order to accommodate the variation in the duration of untrimmed video and class imbalance, we introduce a continuous sampling strategy
\cite{feng2021mist}:

After the feature extractor extracts features $\left \{f_i\right \}^N_{i=1}$ from the original video $V$$=$$\{v_i\}^N_{i=1}$, we uniformly sample $L$ sub-bags $U =\left \{f_{l, t}\right \}^{L, T}_{l=1,t=1}$ from these clips. Specifically, there are $T$ consecutive clips in each sub-bag, as shown in Figure \ref{fig2}. The $T$ is a hyperparameter that needs to be tuned. In detail, in order to make the anomalous score of the anomalous video clips higher than the normal video clips, we adopt deep MIL ranking loss \cite{sultani2018real} to train MLP. The sampled features are entered into the label generator to predict the corresponding anomaly scores $\left \{s_{l, t}\right \}^{L, T}_{l=1, t=1}$. By performing averaging pooling, the anomaly score $S_l$ for each sub-bag is derived:
\begin{equation}
S_l=\frac{1}{T}\sum^T_{t=1} s_{l,t}.
\end{equation}

Then, the anomaly scores of all instances in the positive bags marked as $S^a=\left \{s_i^a\right \}^N_{i=1}$ are min-max normalized to form clip-level labels:

\begin{equation}
{y}^a_i=\frac{{s}^a_i-\min{{S}^a}}{\max{{S}^a}-\min{{S}^a}},i\in[1,N],
\end{equation}
where ${y}^a_i$ between [0,1] is a clip-level label. Finally, the generated labels $ Y^a =\left \{y_i^a\right \}^N_{i=1}$ are mixed with the clip-level labels $Y^n$ to train the feature encoder.

\subsection{The Composition of Feature Encoder}
As shown in Figure \ref{fig3}, we take the feature extractor (i.e. C3D/I3D) as the backbone and introduce two modules: self-guided attention module and spatio-temporal graph module. Compared to a feature encoder with self-guided attention \cite{feng2021mist}, spatio-temporal graph module can help the encoder to capture spatio-temporal contextual anomalies in the features. Furthermore, in order to fully consider the candidate's behavior in the proctoring scenario, we use OpenFace 2.0 \cite{baltrusaitis2018openface} to extract the multi-modal features of student, and comprehensively analyze these features in spatio-temporal graph module.

\subsubsection{Self-Guided Attention Module}
We introduce a self-guided attention module to increase the capabilities of discriminative representation of the feature encoder. To be specific, we extract feature maps $F_4$ and $F_5$ from the fourth and fifth blocks of the base backbone. In addition to this, the encoder introduces two convolutional sub-modules, $M_1$ and $M_2$. Module $M_1$ is responsible for generating the attention map, which is used as the input together with $F_5$ for the attention mechanism. The formula is as follows:

\begin{equation}
A^*=F_5+M_1(F_4)\circ F_5,
\end{equation}
where $\circ$ is the element-level multiplication. $A^*$ finally acts as input to the fully connected layer by global averaging pooling. At the same time, the feature map $F_4$ is converted to $F^*$ after passing through $M_2$. It should be noted that $F^*$ has $2K$ channels as $K$ multiple detectors for the two categories (i.e. normal and abnormal), to enhance the guided supervision \cite{yang2018weakly,feng2021mist}. Then, $F^*$ compresses its channels through spatio-temporal average pooling and $K$ channel-wise average pooling to acquire the guided anomaly score, which is further optimized by $L_2$ to guide the optimization of the feature map produced by $M_1$ and strengthen the attention map $A^*$ generation indirectly.

\subsubsection{Multi-modal Feature Fusion} Generally, factors such as the candidate's facial expression, head posture, eye gaze, body posture, and other factors can be regarded as clues to judge cheating. Thus, we extract features from four perspectives with the help of OpenFace 2.0 and C3D/I3D. OpenFace 2.0 is applied to capture gaze, head pose, and facial action unit features, while body pose and background features are extracted from the C3D/I3D network in the feature encoder.

Eye gaze: By extracting the candidate's eye position through OpenFace 2.0, the gaze direction vector and radian gaze direction can be obtained. Further, We use the standard deviation and maximum range of variation for each dimension of all frames in each clip as a description of the gaze information. In order to analyze the eye status of the candidate in general, We further calculate the time trajectory of the average position of all the eye landmarks for each eye and add it to the feature.

Head pose: We extract the head position of each frame and the head pose vector expressed in radians to describe the head attitude information. The difference between the pose vector and the average position of all frames in each clip is calculated and added to the feature. In addition, we used the average position of 68 facial landmarks to represent the posture state of the head. The standard deviation and maximum variation in the head state of all frames in each clip are calculated and jointly applied with the other head features.

Facial action unit: There are 18 facial key points (1, 2, 4, 5, 6, 7, 9, 10, 12, 14, 15, 17, 20, 23, 25, 26, 28, 45) considered as facial action unit \cite{niu2018automatic}. The intensities of the presence of these action units are generated for further analysis. Firstly, OpenFace 2.0 is used to estimate the intensity of the action unit. Then, we calculate the maximum intensity, maximum variation, and standard deviation of intensity for each action unit in each clip as the feature. In addition, the presence of individual action units is taken into consideration, and the existence frequency for each action unit is calculated and stitched to the final feature.

Body pose and background: The C3D/I3D network was chosen to extract body pose and background features because it could better model the time signal of the input information through 3D convolution and 3D pooling operations, ensuring the timeliness of the input data. As mentioned above, the feature map $A^*$ is generated in the Self-Guided Attention Module and regarded as the input to the fully connected layer by global averaging pooling. We then extract the 128-dimensional features output by the second fully connected layer for feature fusion as body pose and background features.

Finally, the concatenated feature generated by eye gaze, head pose, facial action unit,  body pose, and background is a 245-dimensional vector that can be used to represent the candidate's emotional, cognitive, behavioral, and surrounding situations. A detailed summary of the concatenated feature is shown in Table \ref{Table I.}. The feature is applied for further analysis and we find it very rich and robust to predict cheating behaviors.

\begin{table}[!t]
	\caption{A detailed summary of the concatenated feature.}
	\begin{center}
		\setlength{\tabcolsep}{2mm}{
			\begin{tabular}{c|c|c}
				\midrule
				Feature               & Coded Information       & Dimension \\ \midrule
				\multirow{3}*{ Eye gaze}    & gaze direction vectors      & 12    \\
				 & gaze direction in radians      & 4   \\
				 & 2D \& 3D eye landmarks      & 10    \\ \midrule
				 \multirow{3}*{ Head pose}    & head location      & 6    \\
				 & head pose vector      & 6   \\
				 & 2D \& 3D facial landmarks     & 10    \\ \midrule
				  \multirow{2}*{ Facial action unit}    & presence frequency      & 18    \\
				 &  intensity      & 51   \\ \midrule
				Body pose and background      & 3D feature &  128   \\ \midrule
		\end{tabular}}
		\label{Table I.}
	\end{center}
\end{table}

\subsubsection{Spatio-temporal Graph Module}To be able to exploit the spatio-temporal context in videos, we introduce the spatio-temporal graph module in the feature encoder. The module consists of fully connected layers and GCN with temporal consistency and feature similarity, which is activated by softmax activation function.

Temporal consistency has been pointed out to benefit video-based tasks \cite{zhong2019graph,chang2021contrastive}. Meanwhile, Zhong et al. \cite{zhong2019graph} introduced a temporal consistency graph to solve the anomaly detection problem, proving the effectiveness of temporal consistency in this type of problem. The temporal consistency graph is built on the temporal structure of the video. Assuming that a video consists of N clips, its adjacency matrix $A^T \in R^{N \times N}$ only depends on the temporal positions of the $i^{th}$ th and $j^{th}$ clips. We use a Laplacian kernel as the kernel function to distinguish different temporal distances:
\begin{equation}
A{^T}_{(i,j)}=exp(-||i-j||).
\end{equation}

The nearby vertexes are driven to have the same anomaly label via graph-Laplacian operations. According to the previous work \cite{kipf2016semi}, we further approximate the graph-Laplacian:

\begin{equation}
\hat A^{T}=\widetilde D^{T - \frac{1}{2}}\widetilde A^{T}\widetilde D^{T - \frac{1}{2}},
\label{eq6}
\end{equation}
where the adjacency matrix $\widetilde A^T = A^T + I_n$, and $I_n \in R^{N \times N}$ is the identity matrix; $\widetilde D^T_{(i,i)}= \Sigma_j \widetilde A^T_{(i,j)}$ is the corresponding degree matrix. Therefore, the forward result of the temporal consistency graph  layer is as follows:
\begin{equation}
\hat H^{T}=\sigma(\hat A^T X W),
\end{equation}
where W is a trainable parametric matrix, $\sigma$ is an activation function, and $X \in R^{N \times d}$ represents the input d-dimensional feature of $N$ clips.

Feature similarly is modeled by an attributed graph \cite{zhong2019graph} $F = (V,E,X)$,
where $V$ is the vertex set, $E$ is the edge set, and $X$ is the tribute of vertexes. In detail, $V$ is the video defined in section III, $E$ represents the feature similarity between these segments, and $X \in R^{N \times d}$ represents the d-dimensional features of these $N$ clips. We define adjacency matrix  $A^F \in R^{N \times N}$ of $F$ as:
\begin{equation}
	 A^{F}_{(i,j)}=exp(X_i \cdot X_j - max(X_i \cdot X)).
\end{equation}
where $A^F_{(i,j)}$ measures the feature similarly between the $i^{th}$ and $j^{th}$ clips. In addition, we apply the normalized exponential function to limit the similarity to the range $(0,1]$. According to the graph $F$, clips with similar features are closely connected, and the label assignments are propagated differently in accordance with different adjacency values.

Similarly, we acquire the adjacency matrix $\hat A^F$ as the Equation \ref{eq6} for the graph-Laplacian approximation, and the output of the feature similarity graph layer is calculated as:

\begin{equation}
	\hat H^{F}=\sigma(\hat A^F X W),
\end{equation}
where W is a trainable parametric matrix, $\sigma$ is an activation function, and X represents the input feature matrix of $N$ clips. Finally, the outputs of the above two layers are fused by average pooling and activated by the softmax function to obtain a probability prediction for each vertex in the graph, corresponding to the anomaly probability of the $i^{th}$ clip.

\subsection{Loss Function}
\textbf{Deep MIL Ranking Loss:} In the process of generating labels with the label generator, we hope that the instances in the positive bag have higher anomaly scores than instances in the negative bag. A simple method is using the ranking loss function to encourage positive instances to get high anomaly scores. However, this method requires pre-arrangement of clip-level annotations while MIL only requires video-level ones. Therefore, a MIL ranking objective function \cite{sultani2018real} is proposed. On this basis, we use a sub-bag as a fundamental unit to be compatible with the continuous sampling strategy:
\begin{equation}
\max_{1\le l \le {L}}{f(S^a_l)}>\max_{1\le l \le {L}}{f(S^n_l)},
\end{equation}
where $max$ is the function to find the maximum anomaly score of all sub-bags in each bag. After getting anomaly scores, only the sub-bag with the highest scores in the positive and negative bags are ranked. That is because the positive sub-bag with the highest score is most likely to include the abnormal clips, and the sub-bag with the highest score in the negative bag contains normal clips that are most likely to be misjudged as abnormal. The purpose of our MIL ranking function is to separate positive and negative sub-bags as far as possible in terms of anomaly scores. Therefore, our ranking loss is as follows:
\begin{equation}
l(B^a,B^n)=\max{(0,1-\max_{1\le l \le {L}}{f(S^a_l)}+\max_{1\le l \le {L}}{f(S^n_l)})}.
\end{equation}

In real-life scenarios, abnormal events generally last for a short time. Therefore, the anomaly scores of instances in the positive bag are generally sparse, which indicates that there may be abnormal events in only a few sub-bags. Secondly, anomalous events tend to depend on the context, and anomaly scores should change smoothly between instances. Therefore, we minimize the difference in scores between adjacent sub-bags to highlight context-based abnormal clips. Finally, by introducing sparsity and smoothness constraints to the sub-bag scores, the deep MIL ranking loss is obtained:
\begin{equation}
\begin{aligned}
l(B^a,B^n)=\max{(0,1-\max_{1\le l \le {L}}{f(S^a_l)}+\max_{1\le l \le {L}}{f(S^n_l)})} \\
+ \lambda_1 \underbrace{\sum_{l=1}^{(L-1)}(f(S^a_l) - f(S^a_{l+1}))^2 }_{ \textcircled 1}+ \lambda_2 \underbrace{\sum_{l=1}^L f(S^a_l)}_{\textcircled 2}\label{eq10}
\end{aligned}
\end{equation}
where \textcircled 1 is used as a smooth term, and \textcircled 2 as a sparse term. $\lambda_1$ and $\lambda_2$ are hyperparameters used to balance the sorting loss.

\textbf{Classification Loss:} After gaining the labels $Y^a$ generated by the label generator, we apply the corresponding abnormal video $V^a$, and further combine with $V^n$ and its annotations $Y^n$ to train our feature encoder. We utilize the cross-entropy as the loss function of the feature encoder. Since only a few clips of anomalous videos are anomalous, there exists a class-imbalance problem at training time. We
introduce class-reweighting to cross-entropy loss $L_w$:
\begin{equation}
L_w=-w_0y\log{p} - w_1(1-y)\log{(1-p)},
\end{equation}
where $w_0$ and $w_1$ are class weights for abnormal and normal class. We set up $w_0 = 1.2$ and $w_1 = 0.8$ \cite{feng2021mist}.

\section{Experiments}\label{sec5}
\subsection{Datasets}
We conduct experiments on three real-world surveillance video datasets of different scales: UCF-Crime, ShanghaiTech, and Online Exam Proctoring (OEP).

\textbf{UCF-Crime}  \cite{sultani2018real} is a large-scale real surveillance video dataset. It contains 13 kinds of abnormal categories, a total of 1,900 untrimmed long videos, including 950 abnormal videos and 950 normal videos in the same scene. A total of 1610 videos are adopted for training and 290 videos are utilized for testing.

\textbf{ShanghaiTech} \cite{luo2017revisit} is a medium-scale campus surveillance video dataset containing 437 videos. It includes 130 abnormal events in 13 scenarios. This dataset was first proposed for unsupervised learning, so all abnormal videos are in the testing dataset. In order to adapt to weakly supervised learning, these videos are reorganized into 238 training videos and 199 testing videos \cite{zhong2019graph}.

\textbf{OEP} \cite{atoum2017automated} is an online examination monitoring dataset, composed of 24 different examinees’ surveillance videos. Among them, examinees in the first 15 videos were asked to cheat as much as possible. In order to capture the real exam scene, the last 9 examinees were required to take the real exam. These two requirements enhance the authenticity of the data while enriching the types of cheating methods.

Since the OEP paper uses dual-camera detection, it is able to detect more situations. We remove cheating videos that use second cameras to detect and that don't fit the closed-book exam scenario. Finally, 69 normal videos and 65 abnormal videos are obtained, including $Absence$, $Another Person$, and $Device$. In detail, the $Another Person$ event covers two scenarios: an assistant appears on the screen and the candidate turns his head to communicate with the  Off-screen assistant. The case where the candidate covers his mouth to communicate with the Off-screen helper is not considered, because the cheating behavior would be caught in real-time by the browser's voice monitor. In the closed-book exam, the necessary drawing boards and calculators are given in the electronic form. Therefore, $Device$ event takes into account any prohibited items (which may be carried by the candidate or passed through an assistant), whether they appear on or off the screen. For $Absence$, candidates who leave the monitoring range are sensitively captured. Finally, our model is trained through a random combination of short videos.

\textbf{Evaluation Metrics}: According to previous works\cite{liu2018future,sultani2018real,wan2020weakly}, we exploit the area under the curve (AUC) of
the frame-level receiver operating characteristics (ROC) as the main metric. Higher AUC means stronger detection ability and better performance under different recognition thresholds.

\subsection{Implementation Details}
\par\textbf{Feature Extractor}: In order to verify the universality of our model, we apply two mainstream feature extractors: C3D \cite{tran2015learning} is a 3D convolutional network encoder. We use the pre-trained model on the dataset Sports-1M \cite{karpathy2014large}, and input the features of the layer fc7 into the label generator. Compared to C3D, I3D \cite{carreira2017quo} has a two-stream architecture. We first extract the output $(2 \times 7 \times 7 \times 1024)$ of the mixed\_5c layer in the I3D network, and then use an average pooling 3D with a convolutional kernel $(2 \times 7 \times 7)$ to obtain the 1024-dimensional features as the input of the label generator.

\par \textbf{Label Generator} is a 3-layer MLP with unit numbers of 512, 32, and 1. The dropout probability between each layer is 0.6. In the training process, we regard 16 frames as a clip, and set the hyperparameters: $L=32$, $T=8$, and $\lambda_1=\lambda_2 = 8 \times 10^{-5}$. Finally, we use the Adagrad optimizer to train the generator with a learning rate of 0.01.

\par \textbf{Feature Encoder}: The self-guided attention module consists of 2 encoding units, namely $M_1$, $M_2$. Let $C$ be the channel number of $F_4$ and $2K$ be the channel number of $F^*$. $M_1$ consists of a $3 \times 3 \times 3 \times C$ 3DConv layer with the stride of 2, a $1 \times 1 \times 1 \times 2K$ and a $1 \times 1 \times 1 \times 1$ 3DConv layer. The first two layers are activated by the ReLU function, and the last layer is activated by the Sigmoid; $M_2$ consists of a $3 \times 3 \times 3 \times C$ 3DConv layer with the stride of 2 and two $1 \times 1 \times 1 \times 2K$ 3DConv layers. It is worth noting that $C$ is determined by the number of channels of the feature map derived from the base backbone (C3D/I3D); $K$ is a hyperparameter. Through the hyperparameters analysis in Section \ref{sec5}, it is concluded that the performance of the model reaches best when $K=15$.

In the spatio-temporal graph module, the output dimensions of the first two fully connected layers are 512 and 128, at the 60\% dropout rate. Then, there are two graph convolutional layers applied to each graph: a hidden layer with 32-unit activated by ReLU, and the last 2-unit output layer. After multiple experiments, it was found that the training parameters had minimal impact on performance, so we generally set $base\_learning\_rate = 0.0001 $, $weight\_decay = 0.0005$ and train 300 epochs.

\subsection{Comparison Results}
Compared with several existing works in weakly supervised learning, we evaluate our model CHEESE in terms of accuracy, effectiveness, and universality. In order to adapt MIL to ShanghaiTech and OEP, we re-standardized the datasets. In addition, the applicable scenarios of UCF and ShanghaiTech are different from OEP, and crowded scenes cannot use Openface to extract features, so we removed the feature fusion process in the experiment shown in Table \ref{Table II.} and directly put the 128-dimensional feature into two graphs for learning.

When it comes to the UCF dataset (as shown in Table \ref{Table II.}), The AUC of CHEESE reaches 80.56\%, which is comparable to the performance of other models. These results verify the effectiveness of our proposed method. In addition, compared with the works on the ShanghaiTech dataset, the AUC of CHEESE increases to 90.79\%, which verifies the adaptability of our method on different datasets. Moreover, in both datasets, our method is second only to the method proposed by Feng et al. \cite{feng2021mist} in performance. After analysis, we believe that the excessive noise present in the crowded scene affects the ability of the spatio-temporal graph module to capture the dependence. For OEP datasets, by combining multi-dimensional features in the proctoring scenario, CHEESE can effectively help GCN establish dependencies in the video.

To verify the accuracy and robustness of CHEESE on the OEP dataset, we compare related state-of-the-art unsupervised and weakly supervised learning methods \cite{park2020learning,ionescu2019object,sultani2018real,feng2021mist, zhong2019graph}. In Table \ref{Table III.}, we can find that the AUC of CHEESE reaches 87.58\%, which confirms the accuracy and effectiveness of CHEESE. It is worth mentioning that the method of Ionescu et al. \cite{ionescu2019object} outperforms the unsupervised method proposed by Park et al. \cite{park2020learning} on the OEP dataset. It is concluded that the object detection method extracts key information for cheating behavior detection \cite{ionescu2019object}, which proves that extracting some clues beforehand, such as body posture, head posture, etc., can lead to better performance. Moreover, our method outperforms the method proposed in \cite{feng2021mist} on the OEP dataset. They also used pseudo-labels to optimize the encoder, which shows that the spatio-temporal graph module can effectively improve prediction accuracy. These results validate that our proposed method is more applicable to the proctoring domain than baselines.

\subsection{Ablation Study}

\begin{table}[!t]
	\caption{Abnormal detection results (\%) between existing weakly supervised methods in terms of frame-level AUC on the UCF and the ShanghaiTech datasets.}
	\begin{center}
		\setlength{\tabcolsep}{2.3mm}{
		\begin{tabular}{c|c|cc}
			\midrule
			\multirow{2}{*}{Methods} & \multirow{2}{*}{Feature Encoder}	 & \multicolumn{2}{c}{AUC(\%)}	\\
			&                               & ShanghaiTech      & UCF         \\ \midrule
			SVM                      & C3D                        & -         & 50                    \\
			Sultani et al. \cite{sultani2018real}   & C3D              & 86.30                   & 75.4              \\
			Zhu et al. \cite{zhu2019motion}     & AE                    & -                & 79.0                \\
			Zhong et al. \cite{zhong2019graph}    & C3D                  & 76.44               & 81.08             \\
			\multirow{2}*{AR-Net \cite{wan2020weakly}}  	 & C3D  			 	 & 85.01				 & - 		\\
			  & I3D  	 & 85.38	 & - 		\\
			\multirow{2}*{ MIST \cite{feng2021mist}}		 & C3D  				 & 93.13			 & 81.40  		  \\
					 							 & I3D  				 & \textbf{94.83} 			 & \textbf{82.30}  		 \\ \midrule
			\multirow{2}*{\textbf{CHEESE}}                    & C3D                 & 89.82              & 80.25              \\
		                   							   & I3D                       & 90.79         & 80.56              \\ \midrule
		\end{tabular}}
		\label{Table II.}
	\end{center}
\end{table}

\begin{table}[!t]
	\caption{Abnormal detection results (\%) between different types of methods in terms of frame-level AUC on the OEP datasets.}
	\begin{center}
		\setlength{\tabcolsep}{3mm}{
		\begin{tabular}{c|c|c}
			\midrule
			Methods                & Supervised        & AUC(\%) \\ \midrule
			Park et al. (Recon) \cite{park2020learning}    & Un      & 72.5    \\
			Park et al. (Pred) \cite{park2020learning}  & Un      & 76.3    \\
			Ionescu et al. \cite{ionescu2019object} & Un      & 78.2    \\ \midrule
			Sultani et al. \cite{sultani2018real} & Weak & 81.6    \\
			Zhong et al. \cite{zhong2019graph}   & Weak & 82.36    \\
			MIST \cite{feng2021mist}	& Weak & 83.95 \\   \midrule
		\textbf{CHEESE}                     & Weak & \textbf{87.58}   \\ \midrule
		\end{tabular}}
		\label{Table III.}
	\end{center}
\end{table}

Due to the different scenarios, we cannot utilize Openface to extract the facial features of crowded scenes in the UCF and ShanghaiTech datasets. Therefore, the features and dimensions of the convolutional layers imported into the spatio-temporal graph module are also different. To control variates and take full advantage of multi-modal features, we conduct the following ablation studies on OEP datasets.
\subsubsection{Component Analysis}
We evaluate several variants of our model by eliminating GCN, self-guided attention module, and continuous sampling strategy in CHEESE. To obtain the baseline, the self-guided attention module and spatio-temporal graph module are removed, and the continuous sampling strategy is replaced by the uniform sampling strategy. In order to demonstrate the effectiveness of our GCN structure in the spatio-temporal graph module, we apply fully connected layers to replace graph convolutional layers.  As shown in Table \ref{Table IV.}, we can see that its performance has dropped a lot compared to CHEESE. Next, to demonstrate the performance brought by the self-guided attention module in the encoder, we conduct ablation experiments on CHEESE without the self-guided attention module, and the AUC drop by about 2.8\% from the results. In addition, We apply a uniform sampling strategy instead of the continuous sampling strategy during the label generation phase. Compared to CHEESE, the AUC of the model utilizing a uniform sampling strategy decreased by 1.9\% when $L$ and $T$ are fixed.

\begin{table}[!t]
	\caption{Ablation Studies on OEP dataset. Baseline is the CHEESE applied uniform sampling strategy without two modules. CHEESE is our whole model. {CHEESE}  $^{w/o\ GCN}$, {CHEESE} $^{w/o\ SGA} $, and {CHEESE}  $^{w/o\ CS}$ denote training without GCN in spatio-temporal graph module, continuous sampling strategy, and self-guided attention module, respectively.}
	\begin{center}
		\setlength{\tabcolsep}{6mm}{
			\begin{tabular}{c | c}
				\midrule
				Variants of our model  & AUC   \\ \midrule
				Baseline & 82.93 \\ \midrule
				\textbf{CHEESE}  $^{w/o		\ GCN}$     & 83.36 \\ \midrule
				\textbf{CHEESE}  $^{w/o\ SGA}$     & 84.81 \\ \midrule
				\textbf{CHEESE}  $^{w/o\ CS}$     & 85.66 \\ \midrule
				\textbf{CHEESE}       & \textbf{87.58} \\ \midrule
		\end{tabular}}
		\label{Table IV.}
	\end{center}
\end{table}

\subsubsection{Hyperparameters Analysis}
\begin{figure}[htbp]
	\centering
	\includegraphics[width=0.4\textwidth]{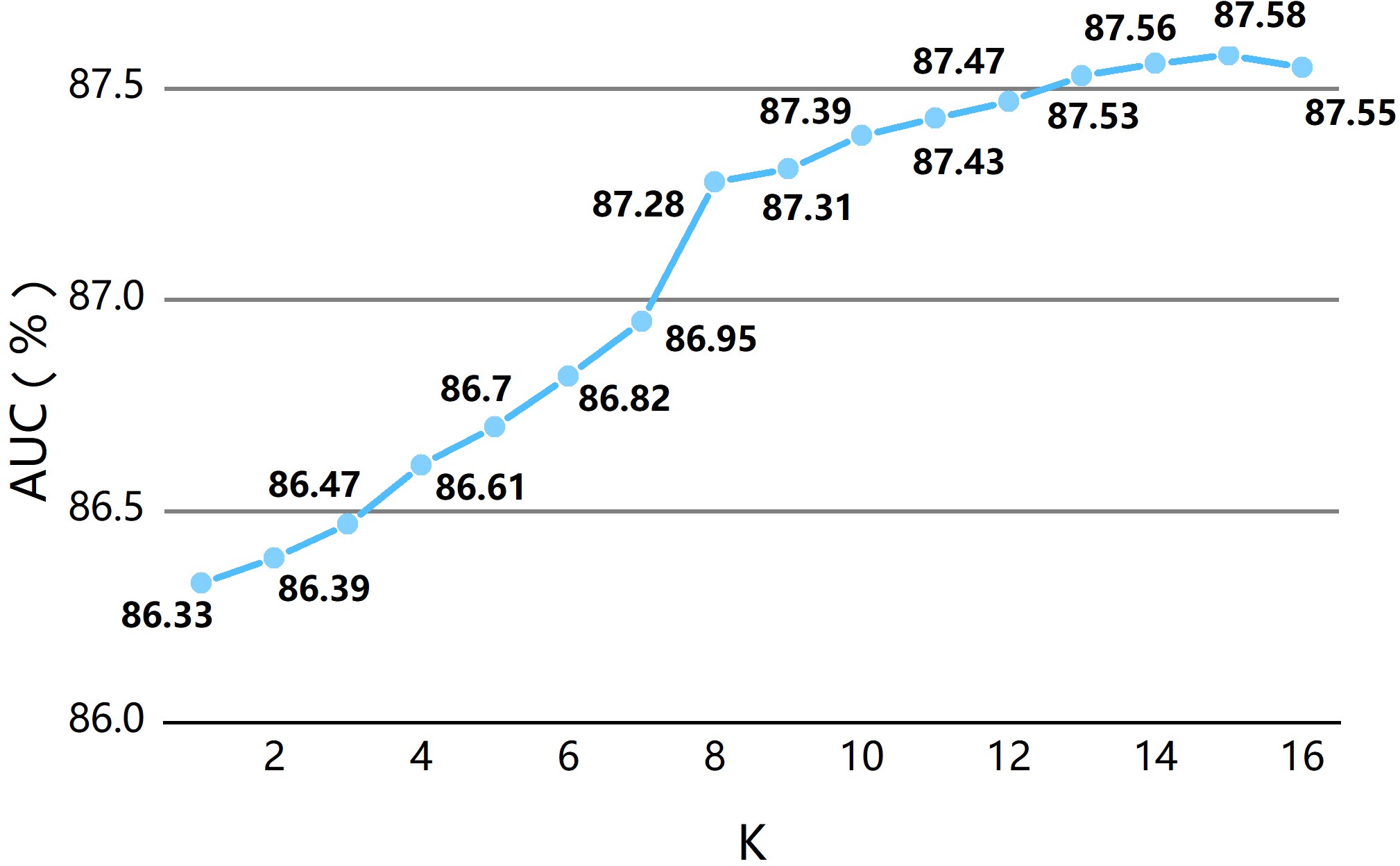}
	\caption{The variations of AUC for different values of the multiple detector $K$ in self-guided attention module on OEP dataset using I3D. }
	\label{fig4}
\end{figure}

\begin{figure}[htbp]
	\centering
	\includegraphics[width=0.4\textwidth]{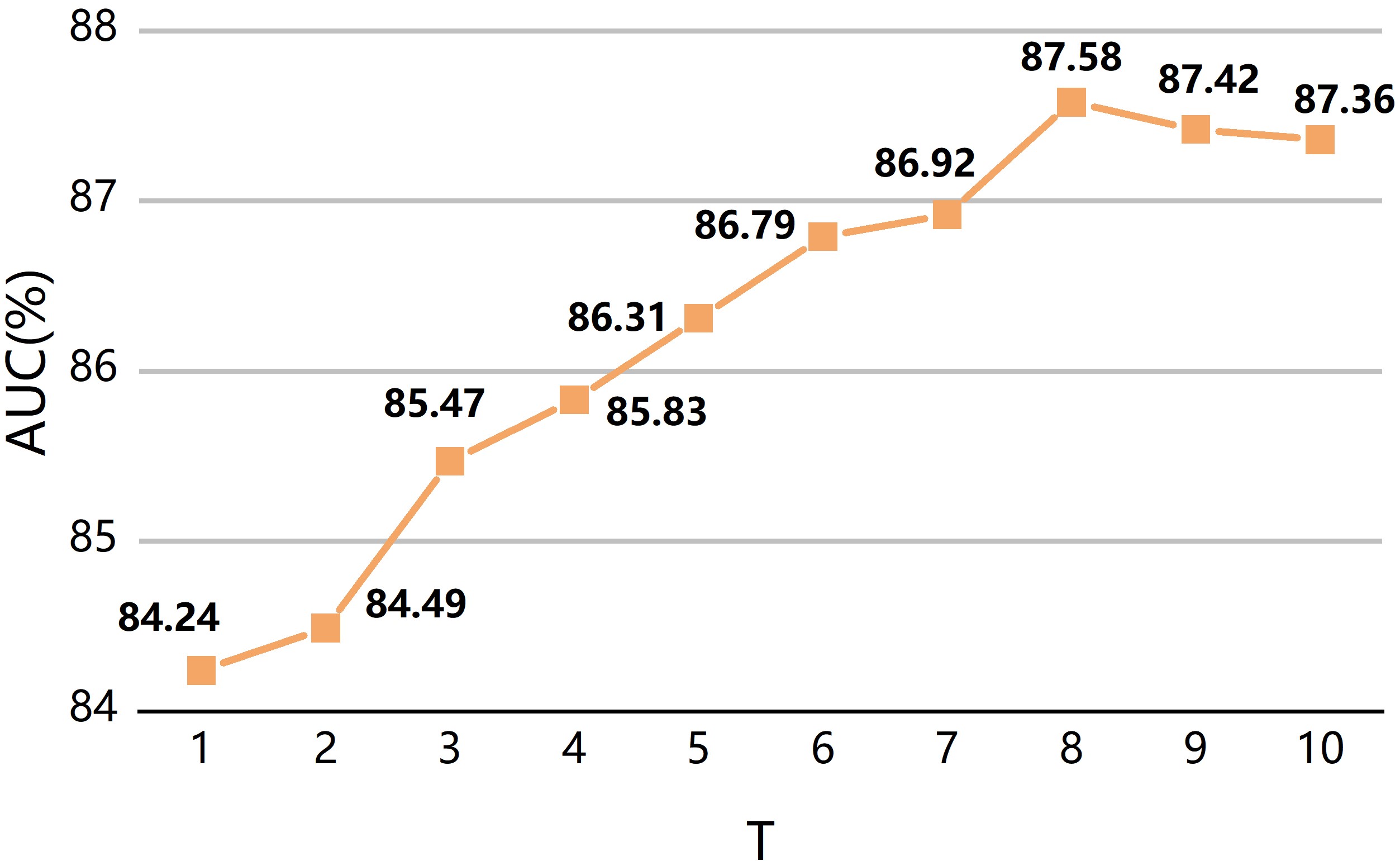}
	\caption{The variations in AUC on OEP dataset using I3D by changing the total number of consecutive instances in a sub-bag $T$. }
	\label{fig5}
\end{figure}

In our model, there are five hyperparameters: the sparsity and smoothness coefficients $\lambda_1$ and $\lambda_2$ in Eq. \ref{eq10}; Multiple detector $K$ in self-guided attention module; The total number $L$ of sub-bags in a bag and the number $T$ of consecutive instances in the sub-bag; To ensure optimal performance, we apply I3D for experiments. Firstly, $\lambda_1$ and $\lambda_2$ are set to $8 \times 10^{-5}$, as Sultani et al. \cite{sultani2018real} proposed.  Considering that over-amplifying the feature maps results in suboptimal performance mainly due to overfitting \cite{yang2018weakly}, We conducted evaluation experiments within $K<16$ (As shown in Figure \ref{fig4}). As $K$ increases, the performance of the model gradually increases and reaches its best when $K=15$. But when $K$ continues to increase, the performance drops instead. Following \cite{sultani2018real, feng2021mist}, we set $L=32$. In addition to this, we investigate the results for different $T$ in Figure \ref{fig5}. From the figure, the impact of the hyperparameter $T$ on the OEP is around 3\%, and the result with $T=8$ outperforms the other results.

\subsection{Runtime Analysis}
We report the runtime of our method for the OEP dataset used C3D in Table \ref{Table V.} with GPU. The results show that current experimental configuration can support real-time videos below 25 FPS, and higher frame rates can lead to issues such as latency or partial frame loss, which can lead to performance degradation. However, the experimental results are not the upper limit of the model's processing speed. In scenarios with high real-time requirements, the processing time of the model can be further reduced by improving the GPU configuration.

Compared to the excellent performance of C3D in real-time processing (processing speed up to \textbf{313} FPS \cite{tran2015learning}), our method spends a lot of time on the {CHEESE}$^{MF}$, which uses Openface 2.0 for face recognition and feature extraction. In other words, the time consumption of CHEESE depends mainly on the runtime of Openface 2.0. Baltrusaitis et al. \cite{baltrusaitis2018openface} stated that the approach they proposed was applicable to real-time environments, which illustrates the feasibility of our approach.

\begin{table}[!t]
	\caption{Runtime analysis on OEP using C3D. {CHEESE}$^{MF}$ represents the multi-modal features fusion stage in the model.}
	\begin{center}
		\setlength{\tabcolsep}{6mm}{
			\begin{tabular}{c | c}
				\midrule
				Variants of our model     &  Average Value of FPS \\ \midrule
				\textbf{CHEESE}$^{MF}$  & 29.2  \\ \midrule
				\textbf{CHEESE} & 25.4    \\ \midrule
		\end{tabular}}
		\label{Table V.}
	\end{center}
\end{table}

\begin{figure}[htbp]
	\centering
	\captionsetup[subfigure]{font= {footnotesize},justification=centering}
	\subfloat[An example of normal video.]{
		\centering
		\includegraphics[width=0.45\textwidth]{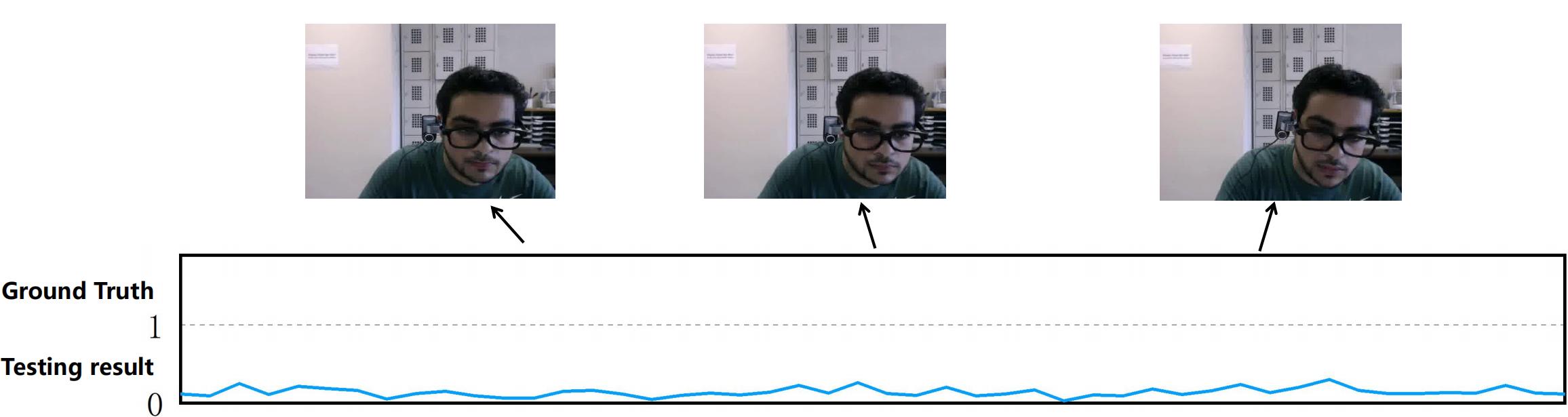}
	}
	\vspace{-4mm}
	\vspace{0mm}
	\subfloat[An example of Another Person event.]{
		\centering
		\includegraphics[width=0.45\textwidth]{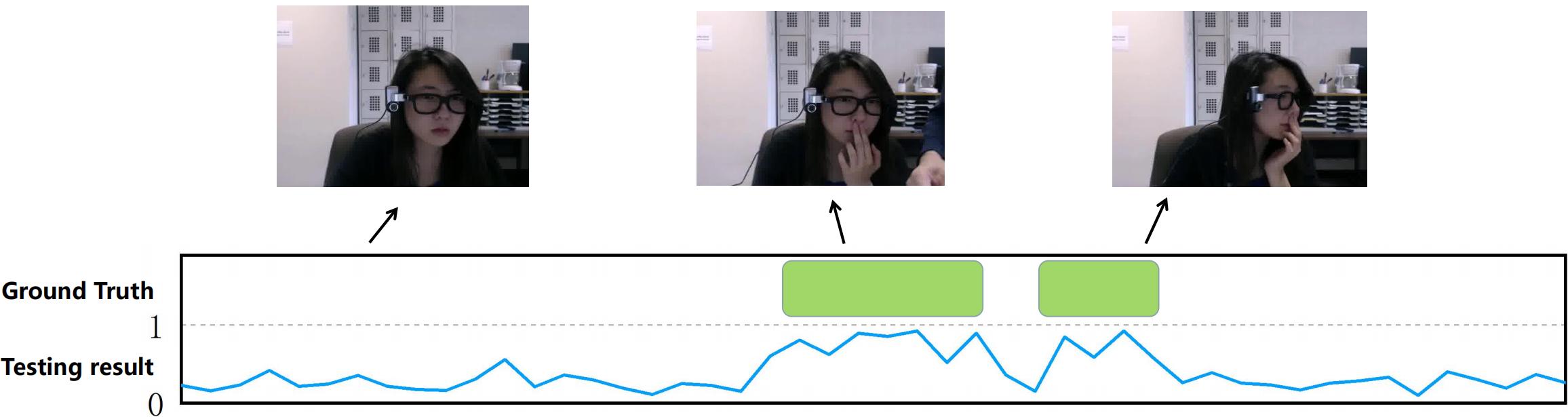}
	}
	\vspace{-4mm}
	\vspace{0mm}
	\subfloat[An example of Device event.]{
		\centering
		\includegraphics[width=0.45\textwidth]{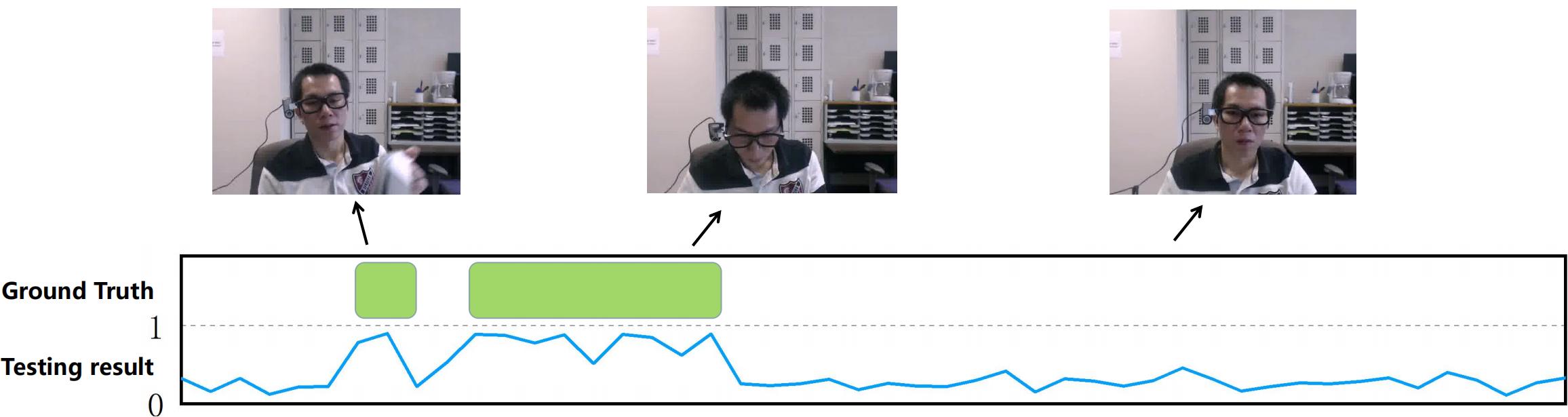}
	}
	\vspace{-4mm}
	\vspace{0mm}
	\subfloat[An example of Absence event.]{
		\centering
		\includegraphics[width=0.45\textwidth]{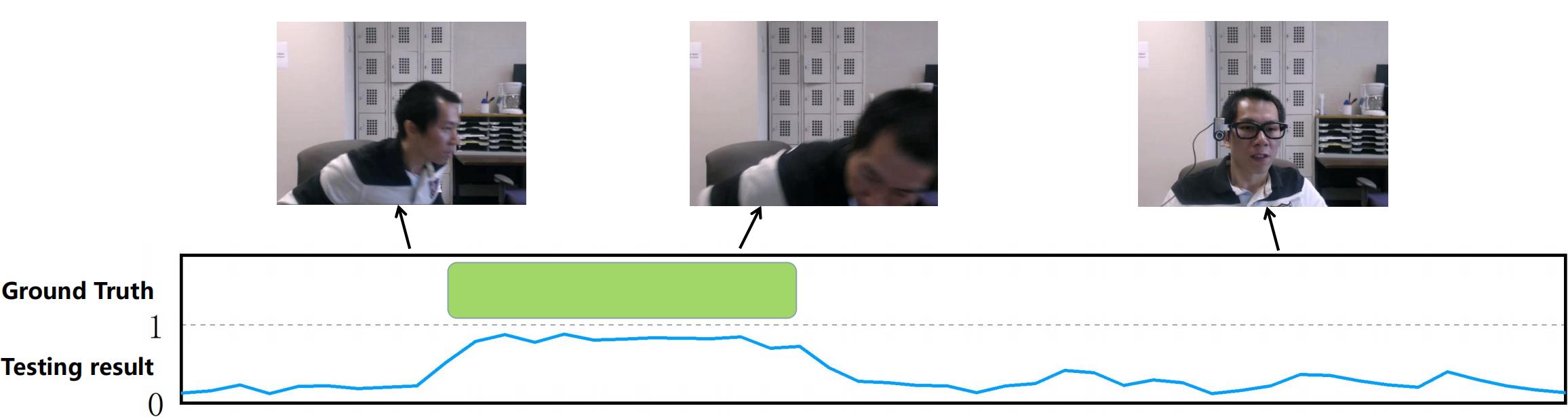}
	}
	\vspace{-4mm}
	\vspace{5mm}
	\caption{Qualitative results of our method on OEP testing videos. The horizontal axes denote the timestamps. (a) is a normal video. (b) (c) (d) are anomalous videos, which contain three types of cheating events. Our model localizes their anomalous events precisely, and generates low anomaly scores. }
	\label{fig6}
\end{figure}

\begin{figure}[htbp]
	\centering
	\subfloat{
		\makebox[7mm][r]{\footnotesize Another Person}
		\begin{minipage}[c]{0.185\linewidth}
			\centering
			\includegraphics[width=1\linewidth]{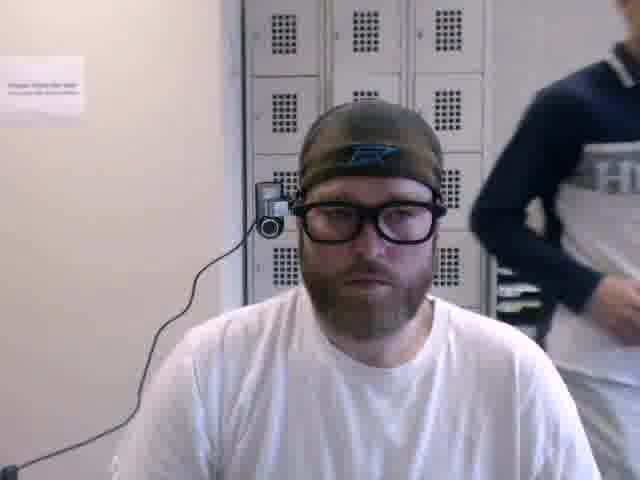}
		\end{minipage}
		\begin{minipage}[c]{0.185\linewidth}
			\centering
			\includegraphics[width=1\linewidth]{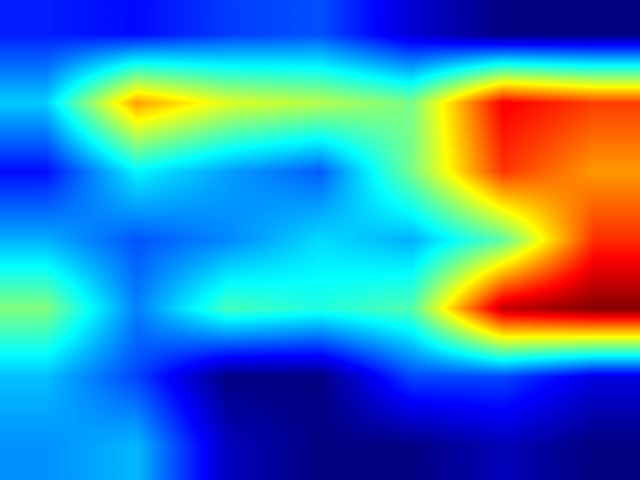}
		\end{minipage}
		\begin{minipage}[c]{0.185\linewidth}
			\centering
			\includegraphics[width=1\linewidth]{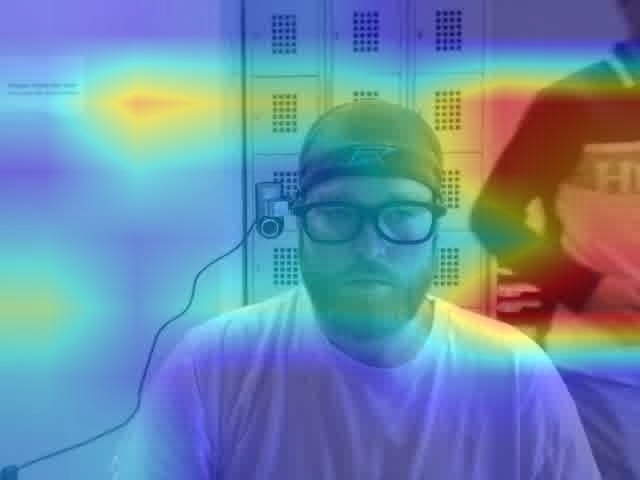}
		\end{minipage}
		
	}
	\vspace{-3mm}
	\quad	
	\vspace{-3mm}
	\subfloat{
		\makebox[7mm][r]{\footnotesize Device}
		\begin{minipage}[c]{0.185\linewidth}
			\centering
			\includegraphics[width=1\linewidth]{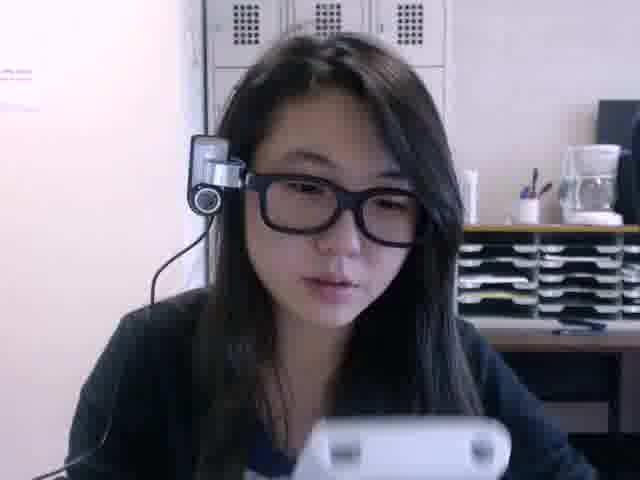}
		\end{minipage}
		\begin{minipage}[c]{0.185\linewidth}
			\centering
			\includegraphics[width=1\linewidth]{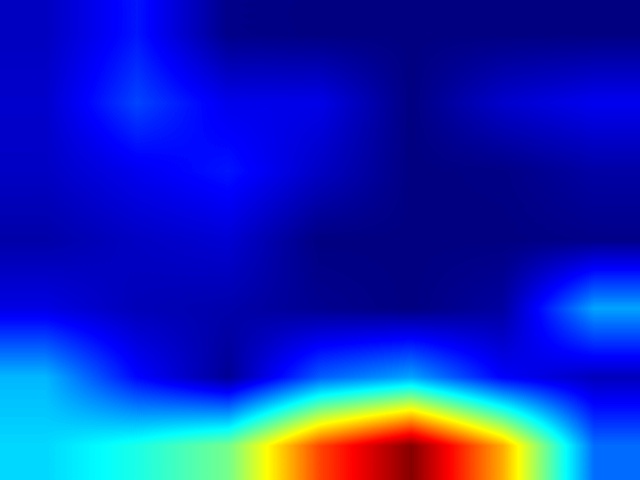}
		\end{minipage}
		\begin{minipage}[c]{0.185\linewidth}
			\centering
			\includegraphics[width=1\linewidth]{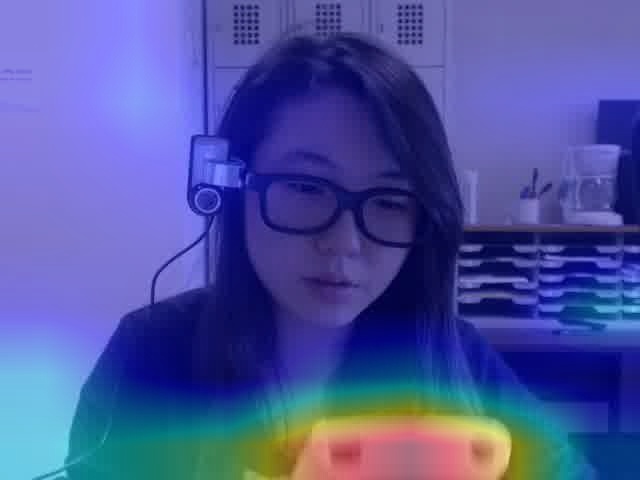}
		\end{minipage}
	}\quad
	\vspace{-3mm}
	\subfloat{
		\makebox[7mm][r]{\footnotesize Absence}
		\begin{minipage}[c]{0.185\linewidth}
			\centering
			\includegraphics[width=1\linewidth]{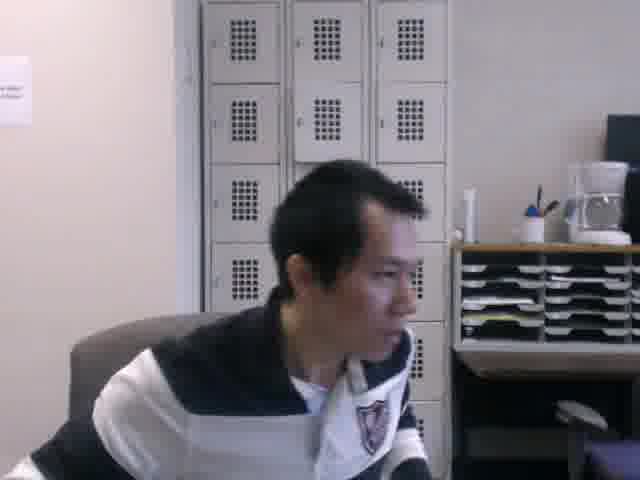}
		\end{minipage}
		\begin{minipage}[c]{0.185\linewidth}
			\centering
			\includegraphics[width=1\linewidth]{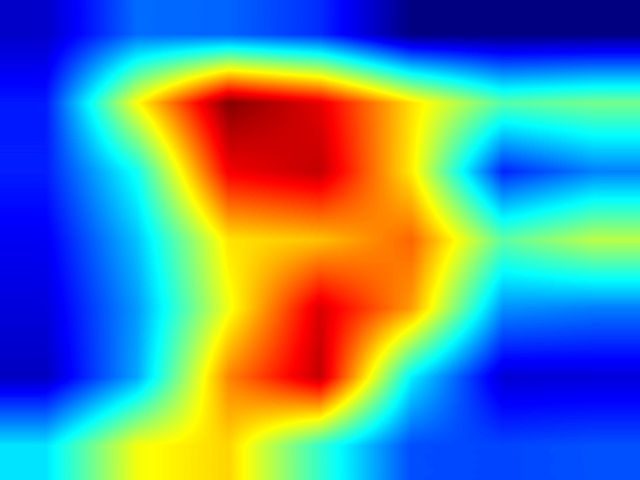}
		\end{minipage}
		\begin{minipage}[c]{0.185\linewidth}
			\centering
			\includegraphics[width=1\linewidth]{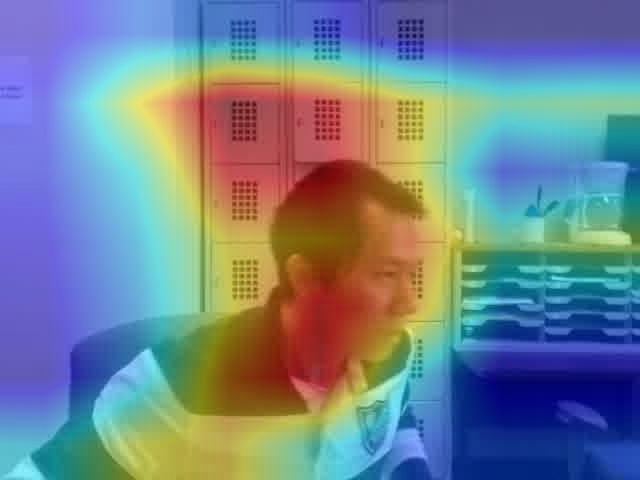}
		\end{minipage}
	}\quad
	\vspace{3mm}
	\subfloat{
		\makebox[7mm][r]{\footnotesize Off-screen assistants}
		\begin{minipage}[c]{0.185\linewidth}
			\centering
			\includegraphics[width=1\linewidth]{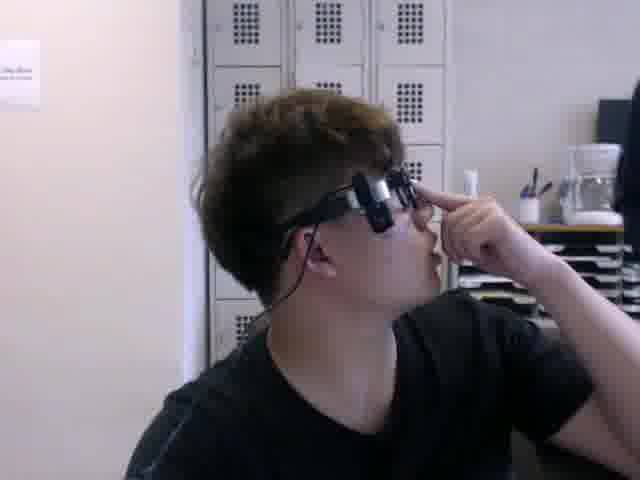}
		\end{minipage}
		\begin{minipage}[c]{0.185\linewidth}
			\centering
			\includegraphics[width=1\linewidth]{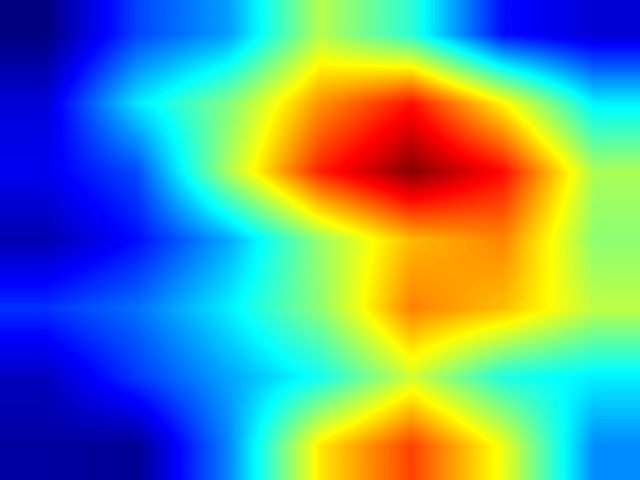}
		\end{minipage}
		\begin{minipage}[c]{0.185\linewidth}
			\centering
			\includegraphics[width=1\linewidth]{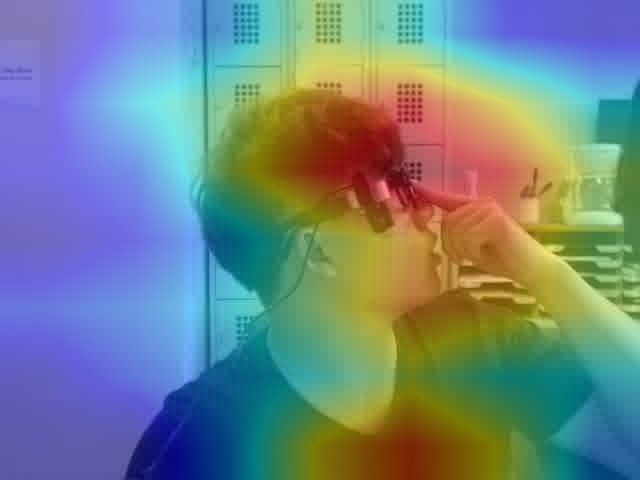}
		\end{minipage}
	}\quad
	\caption{Visualization results of anomaly activation maps on OEP dataset. (better viewed in color).}
	\label{fig7}
\end{figure}

\subsection{Visual Results}
We visualize the qualitative results of CHEESE with 4 videos on OEP dataset. As shown in Figure \ref{fig6}, by learning multi-modal features, CHEESE is able to pinpoint anomalous events and predict anomaly scores very close to zero on normal videos, which demonstrates the effectiveness of our model.

In order to verify the spatial interpretability of the feature encoder, we also visualize the spatial activation map via Grad-CAM \cite{selvaraju2017grad} on $A^*$.
As shown in Figure \ref{fig7}, heat maps of three types of cheating events are collected. When the Another Person event occurs, the attention of CHEESE will turn to the external assistant. Similarly, when devices such as mobile phones, and calculators appear on the screen, the model will continue to focus on the location of these prohibited items. In addition, we try to detect \textit{Another Person} event where the external assistant is not on the screen. In the fourth row of the picture, the student is talking with people outside the screen. By observing the heat map, it can be found that the model will focus on the examinee's eyes and areas surrounding them, and at the same time, it will be judged as anomaly. Above all, CHEESE can sensitively locate the decisive positions that help determine whether the current frame is abnormal, which shows the effectiveness of our proposed method.

\section{Conclusion}\label{sec6}
In this paper, we have proposed a weakly supervised learning method to detect cheating behaviors using multi-modal features and GCN. Experimental results demonstrate that our method CHEESE not only achieves state-of-the-art results on the OEP dataset, but also has competitive performance on other anomaly detection datasets.

In future work, we will continue to improve the performance of CHEESE in three aspects: 1) Collecting more proctored videos to capture more types of cheating behaviors, thereby improving the integrity of the model. 2) Combining video, text, and voice information for acquiring more multi-modal features to detect various categories of cheating events. 3) Finally, how to reduce the noise generated by pseudo labels in the first stage is a very important issue. In the first stage, we filter the predicted labels of normal videos and use a concise MLP network to reduce noise. However, MLP is not the best choice, and designing a new model to replace MLP networks is a major focus of future work.

\section*{Acknowledgments}
The authors would like to thank Computer Vision Lab in Michigan State University for providing the OEP dataset.

\bibliographystyle{IEEEtran}
\bibliography{reference}

\begin{thebibliography}{10}
\providecommand{\url}[1]{#1}
\csname url@samestyle\endcsname
\providecommand{\newblock}{\relax}
\providecommand{\bibinfo}[2]{#2}
\providecommand{\BIBentrySTDinterwordspacing}{\spaceskip=0pt\relax}
\providecommand{\BIBentryALTinterwordstretchfactor}{4}
\providecommand{\BIBentryALTinterwordspacing}{\spaceskip=\fontdimen2\font plus
\BIBentryALTinterwordstretchfactor\fontdimen3\font minus
  \fontdimen4\font\relax}
\providecommand{\BIBforeignlanguage}[2]{{%
\expandafter\ifx\csname l@#1\endcsname\relax
\typeout{** WARNING: IEEEtran.bst: No hyphenation pattern has been}%
\typeout{** loaded for the language `#1'. Using the pattern for}%
\typeout{** the default language instead.}%
\else
\language=\csname l@#1\endcsname
\fi
#2}}
\providecommand{\BIBdecl}{\relax}
\BIBdecl

\bibitem{guo2021educational}
T.~Guo, X.~Bai, X.~Tian, S.~Firmin, and F.~Xia, ``Educational anomaly
  analytics: Features, methods, and challenges,'' \emph{Frontiers in Big Data},
  vol.~4, pp. 811--840, 2022.

\bibitem{liu2018future}
W.~Liu, W.~Luo, D.~Lian, and S.~Gao, ``Future frame prediction for anomaly
  detection – a new baseline,'' in \emph{Proceedings of the IEEE Conference
  on Computer Vision and Pattern Recognition (CVPR)}, 2018, pp. 6536--6545.

\bibitem{ionescu2019object}
R.~T. Ionescu, F.~S. Khan, M.-I. Georgescu, and L.~Shao, ``Object-centric
  auto-encoders and dummy anomalies for abnormal event detection in video,'' in
  \emph{Proceedings of the IEEE/CVF Conference on Computer Vision and Pattern
  Recognition (CVPR)}, 2019, pp. 7842--7851.

\bibitem{tudor2017unmasking}
R.~Tudor~Ionescu, S.~Smeureanu, B.~Alexe, and M.~Popescu, ``Unmasking the
  abnormal events in video,'' in \emph{Proceedings of the IEEE International
  Conference on Computer Vision (ICCV)}, 2017, pp. 2895--2903.

\bibitem{nguyen2019anomaly}
T.-N. Nguyen and J.~Meunier, ``Anomaly detection in video sequence with
  appearance-motion correspondence,'' in \emph{Proceedings of the IEEE/CVF
  International Conference on Computer Vision (ICCV)}, 2019, pp. 1273--1283.

\bibitem{fang2020anomaly}
Z.~Fang, J.~Liang, J.~T. Zhou, Y.~Xiao, and F.~Yang, ``Anomaly detection with
  bidirectional consistency in videos,'' \emph{IEEE Transactions on Neural
  Networks and Learning Systems}, vol.~33, no.~3, pp. 1079--1092, 2022.

\bibitem{lu2013abnormal}
C.~Lu, J.~Shi, and J.~Jia, ``Abnormal event detection at 150 fps in matlab,''
  in \emph{Proceedings of the IEEE International Conference on Computer Vision
  (ICCV)}, 2013, pp. 2720--2727.

\bibitem{zhao2017spatio}
Y.~Zhao, B.~Deng, C.~Shen, Y.~Liu, H.~Lu, and X.~Hua, ``Spatio-temporal
  autoencoder for video anomaly detection,'' in \emph{Proceedings of the 25th
  ACM International Conference on Multimedia}, 2017, pp. 1933--1941.

\bibitem{luo2017revisit}
W.~Luo, W.~Liu, and S.~Gao, ``A revisit of sparse coding based anomaly
  detection in stacked rnn framework,'' in \emph{Proceedings of the IEEE
  International Conference on Computer Vision (ICCV)}, 2017, pp. 341--349.

\bibitem{sultani2018real}
W.~Sultani, C.~Chen, and M.~Shah, ``Real-world anomaly detection in
  surveillance videos,'' in \emph{Proceedings of the IEEE Conference on
  Computer Vision and Pattern Recognition (CVPR)}, 2018, pp. 6479--6488.

\bibitem{gong2019memorizing}
D.~Gong, L.~Liu, V.~Le, B.~Saha, M.~R. Mansour, S.~Venkatesh, and A.~v.~d.
  Hengel, ``Memorizing normality to detect anomaly: Memory-augmented deep
  autoencoder for unsupervised anomaly detection,'' in \emph{Proceedings of the
  IEEE/CVF International Conference on Computer Vision (ICCV)}, 2019, pp.
  1705--1714.

\bibitem{zhou2019anomalynet}
J.~T. Zhou, J.~Du, H.~Zhu, X.~Peng, Y.~Liu, and R.~S.~M. Goh, ``{AnomalyNet:}
  an anomaly detection network for video surveillance,'' \emph{IEEE
  Transactions on Information Forensics and Security}, vol.~14, no.~10, pp.
  2537--2550, 2019.

\bibitem{nguyen2018weakly}
P.~Nguyen, T.~Liu, G.~Prasad, and B.~Han, ``Weakly supervised action
  localization by sparse temporal pooling network,'' in \emph{Proceedings of
  the IEEE Conference on Computer Vision and Pattern Recognition (CVPR)}, 2018,
  pp. 6752--6761.

\bibitem{sun20223d}
N.~Sun, J.~Tao, J.~Liu, H.~Sun, and G.~Han, ``3d facial feature reconstruction
  and learning network for facial expression recognition in the wild,''
  \emph{IEEE Transactions on Cognitive and Developmental Systems}, 2022, doi:
  {10.1109/TCDS.2022.3157772}.

\bibitem{cheung2015eye}
Y.-m. Cheung and Q.~Peng, ``Eye gaze tracking with a web camera in a desktop
  environment,'' \emph{IEEE Transactions on Human-Machine Systems}, vol.~45,
  no.~4, pp. 419--430, 2015.

\bibitem{zhong2019graph}
J.~Zhong, N.~Li, W.~Kong, S.~Liu, T.~H. Li, and G.~Li, ``Graph convolutional
  label noise cleaner: Train a plug-and-play action classifier for anomaly
  detection,'' in \emph{Proceedings of the IEEE/CVF Conference on Computer
  Vision and Pattern Recognition (CVPR)}, 2019, pp. 1237--1246.

\bibitem{chang2021contrastive}
S.~Chang, Y.~Li, S.~Shen, J.~Feng, and Z.~Zhou, ``Contrastive attention for
  video anomaly detection,'' \emph{IEEE Transactions on Multimedia}, vol.~24,
  pp. 4067--4076, 2021.

\bibitem{wang2018videos}
X.~Wang and A.~Gupta, ``Videos as space-time region graphs,'' in
  \emph{Proceedings of the European Conference on Computer Vision (ECCV)},
  2018, pp. 399--417.

\bibitem{markovitz2020graph}
A.~Markovitz, G.~Sharir, I.~Friedman, L.~Zelnik-Manor, and S.~Avidan, ``Graph
  embedded pose clustering for anomaly detection,'' in \emph{Proceedings of the
  IEEE/CVF Conference on Computer Vision and Pattern Recognition (CVPR)}, 2020,
  pp. 10\,539--10\,547.

\bibitem{li2020peer}
H.~Li, H.~Wei, Y.~Wang, Y.~Song, and H.~Qu, ``Peer-inspired student performance
  prediction in interactive online question pools with graph neural network,''
  in \emph{Proceedings of the 29th ACM International Conference on Information
  \& Knowledge Management}, 2020, pp. 2589--2596.

\bibitem{tian2021weakly}
Y.~Tian, G.~Pang, Y.~Chen, R.~Singh, J.~W. Verjans, and G.~Carneiro,
  ``Weakly-supervised video anomaly detection with robust temporal feature
  magnitude learning,'' in \emph{Proceedings of the IEEE/CVF International
  Conference on Computer Vision (ICCV)}, 2021, pp. 4975--4986.

\bibitem{niu2018automatic}
X.~Niu, H.~Han, J.~Zeng, X.~Sun, S.~Shan, Y.~Huang, S.~Yang, and X.~Chen,
  ``Automatic engagement prediction with gap feature,'' in \emph{Proceedings of
  the 20th ACM International Conference on Multimodal Interaction}, 2018, pp.
  599--603.

\bibitem{pang2021deep}
G.~Pang, C.~Shen, L.~Cao, and A.~V.~D. Hengel, ``Deep learning for anomaly
  detection: A review,'' \emph{ACM Computing Surveys (CSUR)}, vol.~54, no.~2,
  pp. 1--38, 2021.

\bibitem{ren2021deep}
J.~Ren, F.~Xia, Y.~Liu, and I.~Lee, ``Deep video anomaly detection:
  Opportunities and challenges,'' in \emph{2021 International Conference on
  Data Mining Workshops (ICDMW)}, 2021, pp. 959--966.

\bibitem{zhao2011online}
B.~Zhao, L.~Fei-Fei, and E.~P. Xing, ``Online detection of unusual events in
  videos via dynamic sparse coding,'' in \emph{Proceedings of the IEEE/CVF
  Conference on Computer Vision and Pattern Recognition (CVPR)}, 2011, pp.
  3313--3320.

\bibitem{zhang2021video}
Z.~Zhang, S.~Zhong, and Y.~Liu, ``Video abnormal event detection via context
  cueing generative adversarial network,'' in \emph{2021 IEEE International
  Conference on Multimedia and Expo (ICME)}, 2021, pp. 1--6.

\bibitem{perera2019ocgan}
P.~Perera, R.~Nallapati, and B.~Xiang, ``{OCGAN:} one-class novelty detection
  using gans with constrained latent representations,'' in \emph{Proceedings of
  the IEEE/CVF Conference on Computer Vision and Pattern Recognition (CVPR)},
  2019, pp. 2898--2906.

\bibitem{zhai2016deep}
S.~Zhai, Y.~Cheng, W.~Lu, and Z.~Zhang, ``Deep structured energy based models
  for anomaly detection,'' in \emph{Proceedings of The 33rd International
  Conference on Machine Learning}, 2016, pp. 1100--1109.

\bibitem{sabokrou2015real}
M.~Sabokrou, M.~Fathy, M.~Hoseini, and R.~Klette, ``Real-time anomaly detection
  and localization in crowded scenes,'' in \emph{Proceedings of the IEEE
  Conference on Computer Vision and Pattern Recognition (CVPR) Workshops},
  2015, pp. 56--62.

\bibitem{pang2020self}
G.~Pang, C.~Yan, C.~Shen, A.~v.~d. Hengel, and X.~Bai, ``Self-trained deep
  ordinal regression for end-to-end video anomaly detection,'' in
  \emph{Proceedings of the IEEE/CVF Conference on Computer Vision and Pattern
  Recognition (CVPR)}, 2020, pp. 12\,173--12\,182.

\bibitem{park2020learning}
H.~Park, J.~Noh, and B.~Ham, ``Learning memory-guided normality for anomaly
  detection,'' in \emph{Proceedings of the IEEE/CVF Conference on Computer
  Vision and Pattern Recognition (CVPR)}, 2020, pp. 14\,372--14\,381.

\bibitem{xu2017detecting}
D.~Xu, Y.~Yan, E.~Ricci, and N.~Sebe, ``Detecting anomalous events in videos by
  learning deep representations of appearance and motion,'' \emph{Computer
  Vision and Image Understanding}, vol. 156, pp. 117--127, 2017.

\bibitem{ruff2018deep}
L.~Ruff, R.~Vandermeulen, N.~Goernitz, L.~Deecke, S.~A. Siddiqui, A.~Binder,
  E.~M{\"u}ller, and M.~Kloft, ``Deep one-class classification,'' in
  \emph{Proceedings of the 35th International Conference on Machine Learning},
  2018, pp. 4393--4402.

\bibitem{costagliola2008monitoring}
G.~Costagliola, V.~Fuccella, M.~Giordano, and G.~Polese, ``Monitoring online
  tests through data visualization,'' \emph{IEEE Transactions on Knowledge and
  Data Engineering}, vol.~21, no.~6, pp. 773--784, 2009.

\bibitem{li2015massive}
X.~Li, K.-m. Chang, Y.~Yuan, and A.~Hauptmann, ``Massive open online proctor:
  Protecting the credibility of moocs certificates,'' in \emph{Proceedings of
  the 18th ACM Conference on Computer Supported Cooperative Work \& Social
  Computing (CSCW)}, 2015, pp. 1129--1137.

\bibitem{atoum2017automated}
Y.~Atoum, L.~Chen, A.~X. Liu, S.~D.~H. Hsu, and X.~Liu, ``Automated online exam
  proctoring,'' \emph{IEEE Transactions on Multimedia}, vol.~19, no.~7, pp.
  1609--1624, 2017.

\bibitem{li2021visual}
H.~Li, M.~Xu, Y.~Wang, H.~Wei, and H.~Qu, ``A visual analytics approach to
  facilitate the proctoring of online exams,'' in \emph{Proceedings of the 2021
  CHI Conference on Human Factors in Computing Systems}, 2021, pp. 1--17.

\bibitem{yang2019fsa}
T.~Yang, Y.~Chen, Y.~Lin, and Y.~Chuang, ``{FSA-Net:} learning fine-grained
  structure aggregation for head pose estimation from a single image,'' in
  \emph{Proceedings of the IEEE/CVF Conference on Computer Vision and Pattern
  Recognition (CVPR)}, 2019, pp. 1087--1096.

\bibitem{kuhnke2019deep}
F.~Kuhnke and J.~Ostermann, ``Deep head pose estimation using synthetic images
  and partial adversarial domain adaption for continuous label spaces,'' in
  \emph{Proceedings of the IEEE/CVF International Conference on Computer Vision
  (ICCV)}, 2019, pp. 10\,164--10\,173.

\bibitem{ruiz2018fine}
N.~Ruiz, E.~Chong, and J.~M. Rehg, ``Fine-grained head pose estimation without
  keypoints,'' in \emph{Proceedings of the IEEE Conference on Computer Vision
  and Pattern Recognition (CVPR) Workshops}, 2018, pp. 2074--2083.

\bibitem{wan2020weakly}
B.~Wan, Y.~Fang, X.~Xia, and J.~Mei, ``Weakly supervised video anomaly
  detection via center-guided discriminative learning,'' in \emph{2020 IEEE
  International Conference on Multimedia and Expo (ICME)}, 2020, pp. 1--6.

\bibitem{zhu2019motion}
Y.~Zhu and S.~Newsam, ``Motion-aware feature for improved video anomaly
  detection,'' \emph{arXiv preprint arXiv:1907.10211}, 2019.

\bibitem{feng2021mist}
J.~Feng, F.~Hong, and W.~Zheng, ``{MIST}: Multiple instance self-training
  framework for video anomaly detection,'' in \emph{Proceedings of the IEEE/CVF
  Conference on Computer Vision and Pattern Recognition (CVPR)}, 2021, pp.
  14\,009--14\,018.

\bibitem{kipf2016semi}
T.~N. Kipf and M.~Welling, ``Semi-supervised classification with graph
  convolutional networks,'' in \emph{International Conference on Learning
  Representations}, 2017, pp. 1--14.

\bibitem{ou2016asymmetric}
M.~Ou, P.~Cui, J.~Pei, Z.~Zhang, and W.~Zhu, ``Asymmetric transitivity
  preserving graph embedding,'' in \emph{Proceedings of the 22nd ACM SIGKDD
  International Conference on Knowledge Discovery and Data Mining}, 2016, pp.
  1105--1114.

\bibitem{pfeiffer2014attributed}
J.~J. Pfeiffer~III, S.~Moreno, T.~La~Fond, J.~Neville, and B.~Gallagher,
  ``Attributed graph models: Modeling network structure with correlated
  attributes,'' in \emph{Proceedings of the 23rd International Conference on
  World Wide Web}, 2014, pp. 831--842.

\bibitem{liu2022deep}
J.~Liu, F.~Xia, X.~Feng, J.~Ren, and H.~Liu, ``Deep graph learning for
  anomalous citation detection,'' \emph{IEEE Transactions on Neural Networks
  and Learning Systems}, vol.~33, no.~6, pp. 2543 -- 2557, 2022,
  doi:{10.1109/TNNLS.2022.3145092}.

\bibitem{xia2021graph}
F.~Xia, K.~Sun, S.~Yu, A.~Aziz, L.~Wan, S.~Pan, and H.~Liu, ``Graph learning: A
  survey,'' \emph{IEEE Transactions on Artificial Intelligence}, vol.~2, no.~2,
  pp. 109--127, 2021.

\bibitem{tran2015learning}
D.~Tran, L.~Bourdev, R.~Fergus, L.~Torresani, and M.~Paluri, ``Learning
  spatiotemporal features with 3d convolutional networks,'' in
  \emph{Proceedings of the IEEE International Conference on Computer Vision
  (ICCV)}, 2015, pp. 4489--4497.

\bibitem{karpathy2014large}
A.~Karpathy, G.~Toderici, S.~Shetty, T.~Leung, R.~Sukthankar, and L.~Fei-Fei,
  ``Large-scale video classification with convolutional neural networks,'' in
  \emph{Proceedings of the IEEE Conference on Computer Vision and Pattern
  Recognition (CVPR)}, 2014, pp. 1725--1732.

\bibitem{carreira2017quo}
J.~Carreira and A.~Zisserman, ``Quo vadis, action recognition? a new model and
  the kinetics dataset,'' in \emph{Proceedings of the IEEE Conference on
  Computer Vision and Pattern Recognition (CVPR)}, 2017, pp. 6299--6308.

\bibitem{baltrusaitis2018openface}
T.~Baltrusaitis, A.~Zadeh, Y.~C. Lim, and L.-P. Morency, ``Openface 2.0: Facial
  behavior analysis toolkit,'' in \emph{2018 13th IEEE international conference
  on automatic face \& gesture recognition (FG 2018)}.\hskip 1em plus 0.5em
  minus 0.4em\relax IEEE, 2018, pp. 59--66.

\bibitem{yang2018weakly}
J.~Yang, D.~She, Y.-K. Lai, P.~L. Rosin, and M.-H. Yang, ``Weakly supervised
  coupled networks for visual sentiment analysis,'' in \emph{Proceedings of the
  IEEE Conference on Computer Vision and Pattern Recognition (CVPR)}, 2018, pp.
  7584--7592.

\bibitem{selvaraju2017grad}
R.~R. Selvaraju, M.~Cogswell, A.~Das, R.~Vedantam, D.~Parikh, and D.~Batra,
  ``{Grad-CAM:} visual explanations from deep networks via gradient-based
  localization,'' in \emph{Proceedings of the IEEE International Conference on
  Computer Vision (ICCV)}, 2017, pp. 618--626.

\end{thebibliography}

\begin{IEEEbiography}[{\includegraphics[width=1in,height=1.25in,clip,keepaspectratio]{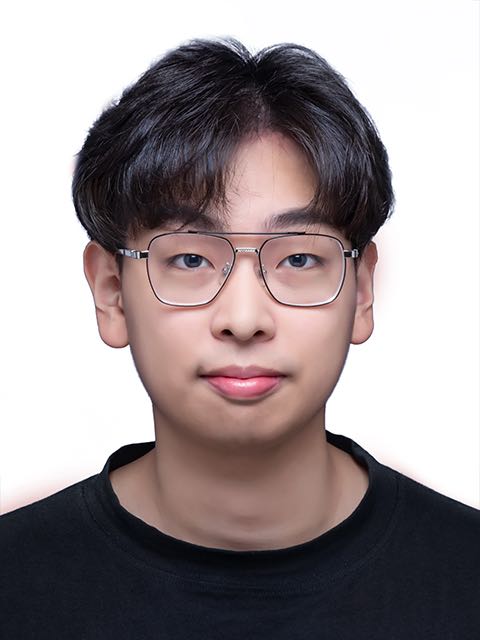}}]{Yemeng Liu}
Yemeng Liu received the B.S. degree from Xinjiang University, China, in 2020. He is currently pursuing the M.S. degree in the School of Software, Dalian University of Technology, China. His research interests include graph learning, computer vision, and anomaly detection.
\end{IEEEbiography}

\begin{IEEEbiography}[{\includegraphics[width=1in,height=1.25in,clip,keepaspectratio]{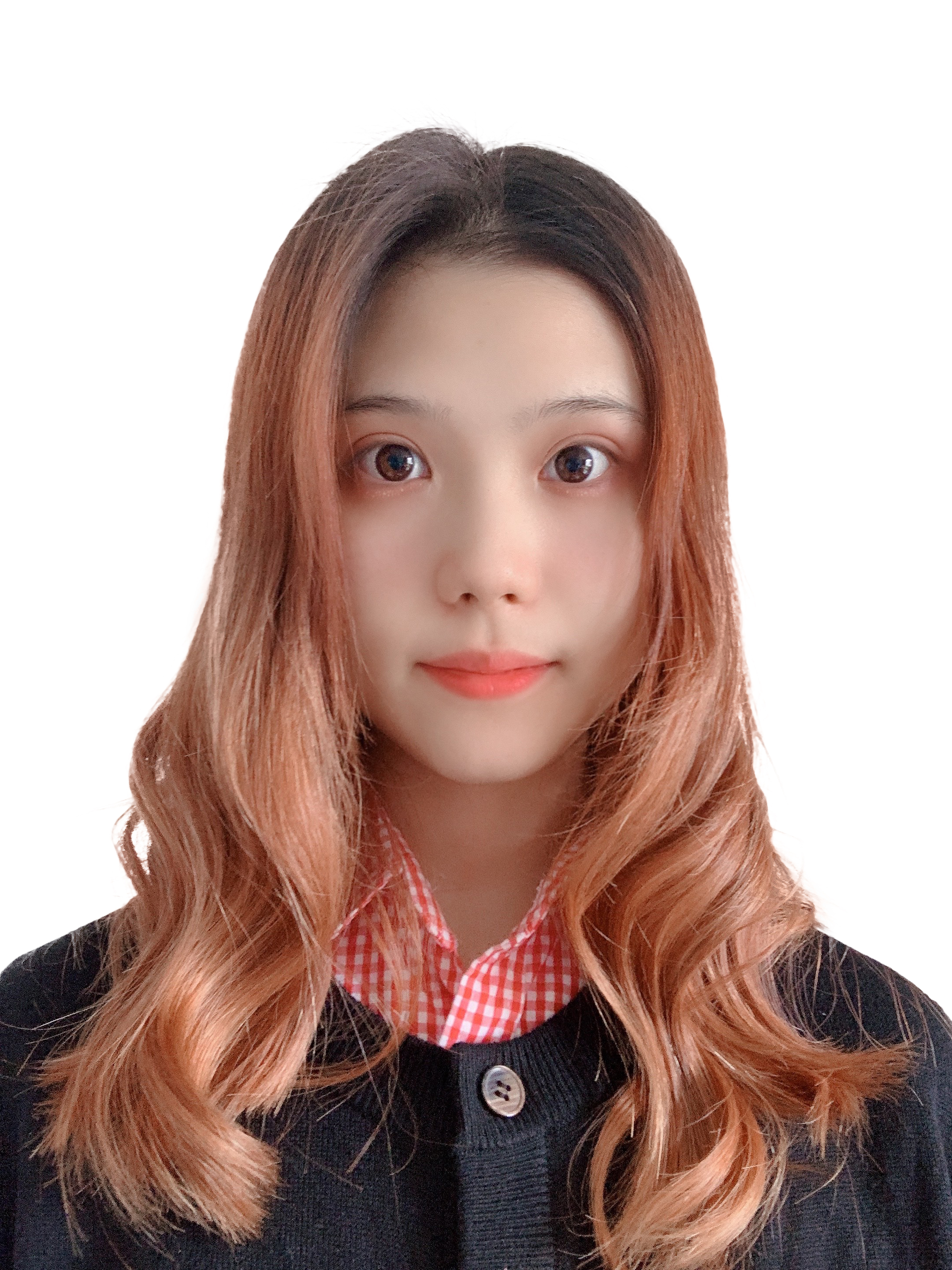}}]{Jing Ren} received the Bachelor degree from Huaqiao University, China, in 2018, and the Master degree from Dalian University of Technology, China, in 2020. She is currently pursuing the Ph.D. degree in the School of Computing Technologies, RMIT University, Melbourne, Australia. Her research interests include data science, graph learning, anomaly detection, and social computing.
\end{IEEEbiography}

\begin{IEEEbiography}[{\includegraphics[width=1in,height=1.25in,clip,keepaspectratio]{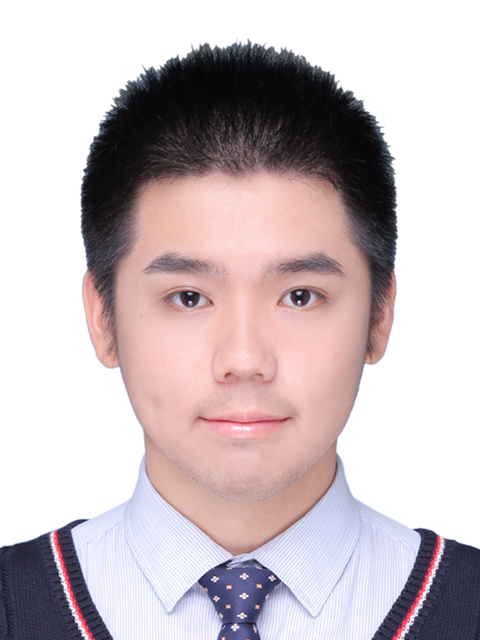}}]{Jianshuo Xu}
received the B.S. degree in School of Software, Dalian University of Technology, China. Currently, he is pursuing the M.S. degree in the Software Engineering Institute, East China Normal University, China. His current research interests include image/video processing, computer vision and formal verification.
\end{IEEEbiography}

\begin{IEEEbiography}[{\includegraphics[width=1in,height=1.25in,clip,keepaspectratio]{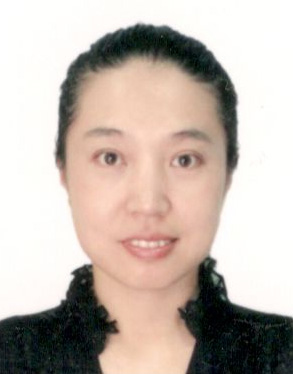}}] {Xiaomei Bai} received the B.Sc. degree from the University of Science and Technology, Liaoning, Anshan, China, in 2000, the M.Sc. degree from Jilin University, Changchun, China, in 2006, and the Ph.D. degree from the School of Software, Dalian University of Technology, Dalian, China, in 2017. Since 2000, she has been working at Anshan Normal University, China. Her research interests include computational social science, science of success in science, and big data.
\end{IEEEbiography}

\begin{IEEEbiography}[{\includegraphics[width=1in,height=1.25in,clip,keepaspectratio]{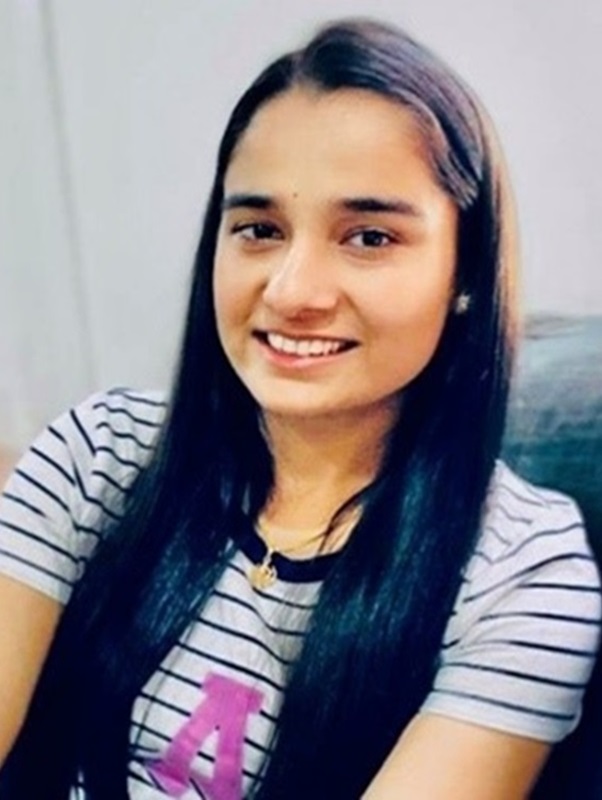}}]{Roopdeep Kaur} received B.Tech degree in Electronics and Communication from Punjab Technical University, India, in 2016 and M.E. degree in Electronics and Communication from Thapar University, India, in 2018. She is currently pursuing a Ph.D. degree in the Institute of Innovation, Science and Sustainability, Federation University Australia. Her current research includes the internet of things, image processing, and data analytics.
\end{IEEEbiography}

\begin{IEEEbiography}[{\includegraphics[width=1in,height=1.25in,clip,keepaspectratio]{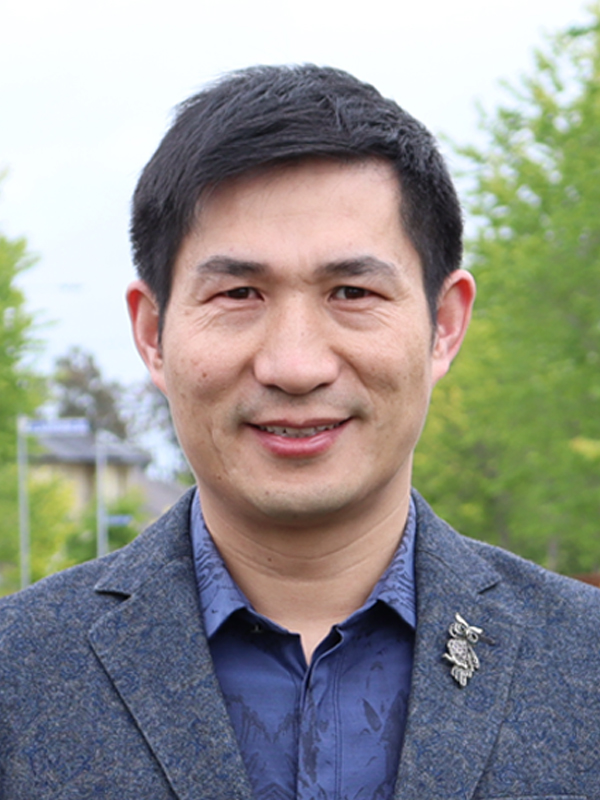}}] {Feng Xia} (Senior Member, IEEE) received the BSc and PhD degrees from Zhejiang University, Hangzhou, China. He is a Professor in School of Computing Technologies, RMIT University, Australia. Dr. Xia has published over 300 scientific papers in international journals and conferences. His research interests include data science, artificial intelligence, graph learning, digital health, and systems engineering. He is a Senior Member of IEEE and ACM, and an ACM Distinguished Speaker.
\end{IEEEbiography}

\vfill

\end{document}